%% file: A-main.tex
\documentclass[a4paper]{cas-dc}

\usepackage[numbers]{natbib}
\usepackage{enumitem}
\usepackage{subcaption} 
\usepackage{amsmath}  
\usepackage[ruled,vlined]{algorithm2e} 
\newtheorem{definition}{Definition} 

\usepackage{enumitem}
\usepackage[normalem]{ulem}

\begin{document}
\let\WriteBookmarks\relax
\def\floatpagepagefraction{1}
\def\textpagefraction{.001}
\let\printorcid\relax        

\newcommand{\ours}{B4}
\newcommand{\ourslong}{\emph{Bias to Behavior from Bull-Bear Dynamics model}}
\newcommand{\Yin}[1]{\textcolor{red}{#1}} 
\newcommand{\edit}[1]{\textcolor{blue}{#1}}

\shorttitle{}    

\shortauthors{}  

\title [mode = title]{From Bias to Behavior: Learning Bull-Bear Market Dynamics with Contrastive Modeling}

\author[1]{Xiaotong Luo}
\ead{xiaotong01@stu2020.jnu.edu.cn}

\author[2]{Shengda Zhuo}
\ead{zhuosd96@gmail.com}

\author[3]{Min Chen}
\ead{minchen2012@hust.edu.cn}

\author[1]{Lichun Li}
\ead{lichun@stu2024.jnu.edu.cn}

\author[1]{Ruizhao Lu}
\ead{a13302600808@stu2022.jnu.edu.cn}

\author[4]{Wenqi Fan}
\ead{wangjieqiu@buaa.edu.cn}

\author[2]{Shuqiang Huang}
\ead{hsq@jnu.edu.cn}

\author[1]{Yin Tang}
\cormark[1] 
\ead{ytang@jnu.edu.cn}

\affiliation[1]{organization={Experimental Center for Economics and Management, Jinan University, Jinan University},
            addressline={Guangzhou 511443},
            state={Guangdong},
            country={China}
            }

\affiliation[2]{organization={College of Cyber Security, Jinan University},
            addressline={Guangzhou 511443},
            state={Guangdong},
            country={China}
            }

\affiliation[3]{organization={School of Computer Science and Engineering, South China University of Technology},
            addressline={Guangzhou 510006}, 
            state={Guangdong},
            country={China}
            }

\affiliation[4]{organization={Department of Computing, Hong Kong Polytechnic University},
            addressline={HongKong 999077}, 
            state={HongKong},
            country={China}
            }

\cortext[1]{Corresponding Authors: Yin Tang}

\newcommand{\alg}{{B4}} 

\begin{abstract}
Financial markets exhibit highly dynamic and complex behaviors shaped by both historical price trajectories and exogenous narratives, such as news, policy interpretations, and social media sentiment.
The heterogeneity in these data and the diverse insight of investors introduce biases that complicate the modeling of market dynamics.
Unlike prior work, this paper explores the potential of bull and bear regimes in investor-driven market dynamics. 
Through empirical analysis on real-world financial datasets, we uncover a dynamic relationship between bias variation and behavioral adaptation, which enhances trend prediction under evolving market conditions.
To model this mechanism, we propose the \ourslong~(\ours), a unified framework that jointly embeds temporal price sequences and external contextual signals into a shared latent space where opposing bull and bear forces naturally emerge, forming the foundation for bias representation. 
Within this space, an inertial pairing module pairs temporally adjacent samples to preserve momentum, while the dual competition mechanism contrasts bullish and bearish embeddings to capture behavioral divergence. 
Together, these components allow \ours~to model bias‑driven asymmetry, behavioral inertia, and market heterogeneity.
Experimental results on real-world financial datasets demonstrate that our model not only achieves superior performance in predicting market trends but also provides interpretable insights into the interplay of biases, investor behaviors, and market dynamics.
\end{abstract}


\maketitle

\input{B-1-Introduction}

\input{B-2-Related.tex}
\input{B-4-Model}
\input{B-5-1-Experiment}

\input{B-5-2-Experiment-Analysis}

\input{B-6-Discussion}
\input{B-7-Conclusion}

\section*{Acknowledgements}
This work was supported by Jinan University, 
the National Natural Science Foundation of China under Grant 62272198, 
Guangdong Key Laboratory of Data Security and Privacy Preserving under Grant 2023B1212060036, 
Guangdong-Hong Kong Joint Laboratory for Data Security and Privacy Preserving under Grant 2023B1212120007, 
Guangdong Basic and Applied Basic Research Foundation under Grant No. 2024A1515010121,
the National Key R\&D Program of China under Grant 2022ZD0116800
the Special Funds for the Cultivation of Guangdong College Students’ Scientific and Technological Innovation (Climbing Program Special Funds) under Grant pdjh2025ak028. 

\printcredits

\bibliographystyle{cas-model2-names}

\bibliography{Ref}

\clearpage
\appendix
\section*{Appendix} 

\input{D-Appendix/D-8-Appendix-Variables}
\input{D-Appendix/D-2-Appendix-Datasets}

\input{D-Appendix/D-7-Appendix-Algorithm}

\input{D-Appendix/D-6-Appendix-Results}

\end{document}

%% file: B-1-Introduction.tex
\section{Introduction}
\label{sec:introduction}

Financial markets are dynamic and adaptive systems, shaped by a complex interplay between historical price trajectories and exogenous factors such as news, policy interpretations, and social media sentiment. 
These factors give rise to cognitive biases that subsequently exert a significant influence on investor behavior, exacerbating the market heterogeneity \cite{ainia2019influence, chang2000examination}. 
Understanding and modeling these dynamics presents a core challenge in finance, necessitating a framework that bridges quantitative and behavioral finance to uncover the intricate relationships between market trends, external influences, and investor decision-making \cite{jin2008behavioral}.

\begin{figure}[!t]
	\centering
	\begin{subfigure}[t]{0.49\linewidth}
		\includegraphics[width=\textwidth]{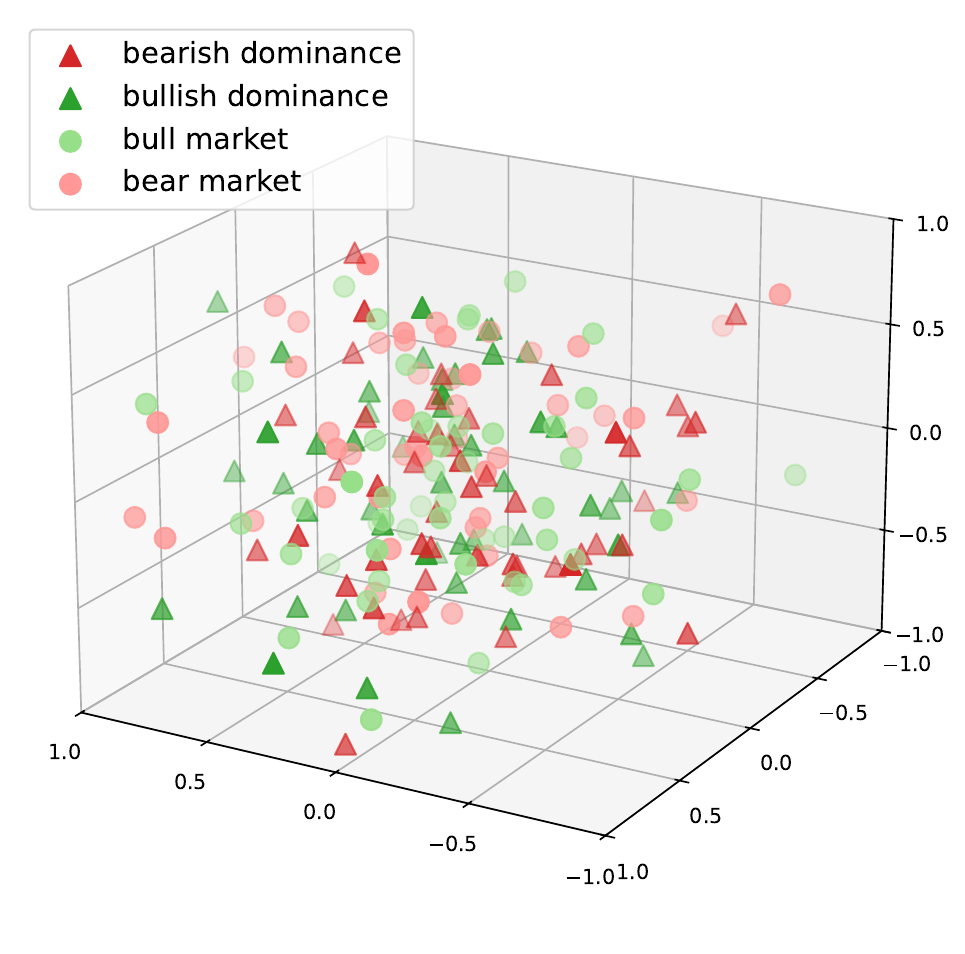}
		\caption{Before \ours}
            \vspace{0.1cm}
		\label{fig:k_means}
	\end{subfigure}
	\begin{subfigure}[t]{0.49\linewidth}
		\includegraphics[width=\textwidth]{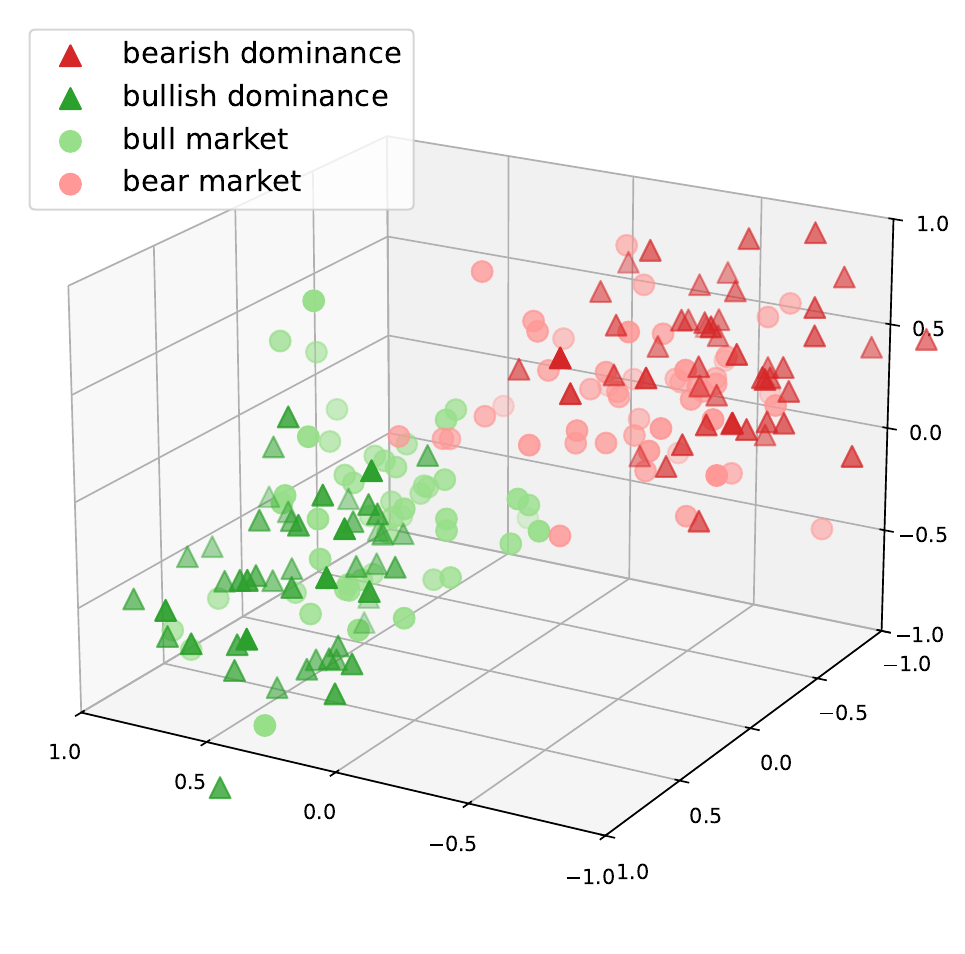}
		\caption{After \ours}
		\label{fig:weighted_k_means}
	\end{subfigure}
	\caption{
        Light green triangles represent ``bullish dominance'' scenarios, and light red triangles represent ``bearish dominance'' scenarios, indicating the competitive interactions between bears and bulls. Red circles denote ``bear markets'', while green circles denote ``bull markets'', reflecting the outcomes of this competition corresponding to downward and upward trends, respectively. In postprocessing of \ours, data points are more clearly differentiated based on market competition dynamics and outcomes.
	}
    \label{fig:BSAP}
\end{figure}

Exploring the mechanisms behind financial market fluctuations has long been a central objective in both economics and quantitative finance.
Early research primarily focused on modeling prices as stochastic processes, employing methods such as random walks \cite{godfrey1964random}, GARCH models \cite{bollerslev1986generalized}, and the Black-Scholes framework \cite{black1973pricing} to describe price randomness and volatility characteristics. 
While these models have been effective in capturing general patterns of price fluctuations, they fall short in capturing ``how'' external narratives are cognitively processed and lead to ``price momentum or reversal'', lacking an explicit mechanism linking information flow to bias formation. This gap has motivated a wave of multimodal and sentiment-aware preditors, such as news \cite{patil2024sentiment} and social media \cite{derakhshan2019sentiment}, gained increasing attention \cite{li2024finreport}. 
While such models boost short-horizon forecasts, they still treat sentiment as an auxiliary signal and overlook the behavioral asymmetry between heterogeneous investor groups. As a result, they fail to explain why the same information may trigger opposing market actions. In essence, few models explicitly model market participants as two sides——bulls and bears——with competing cognitive interpretations and decision inertia. Without that link, predictions remain fragile when dominant sentiment regimes and bias reverses. We summarise two key challenges (\textbf{CHs}):

\par\noindent
\textbf{CH-1: Modality Entanglement and Behavioral Alignment.}
Market dynamics emerge from the interaction of sequential numerical data (\emph{e.g.}, price, volume) and high-dimensional, event-driven textual information. 
These modalities differ not only in format and frequency, but also in interpretability. 
Naive fusion approaches \cite{CHENG2022108218, zhu2023multimodal} obscure temporal causality and fail to model how external narratives are mapped into investor expectations and behaviors, especially under conflicting sentiment conditions. 
Thus, aligning cross-modal signals with investor behavior is critical.

\par\noindent
\textbf{CH-2: Modeling the Bias–Behavior–Price Pathway.} 
Investor reactions are not homogeneous. Bullish and bearish groups often form divergent interpretations of the same event, leading to dynamics shaped by competitive behavior. Yet, most existing works~\cite{liu2024autotimes,ding2024tradexpert} either average out this heterogeneity or treat sentiment as a static attribute. 
To model price movement as the outcome of evolving bias and behavioral inertia, we must capture how attention migrates, how momentum sustains belief, and how conflicting views interact. 
This calls for a framework that tracks how bias initiates, behavior amplifies, and price reacts.

\par\noindent
\textbf{Our work:} 
By quantitatively modeling the impact of cognitive biases and irrational behavior on market price fluctuations, we propose the \ourslong (named \ours), which integrates time-series data with semantic insights from external factors and applies momentum-aware strategies, capable of capturing the competitive dynamics between market biases, sentiment, and market forces.

This paper delivers specific contributions, summarized as follows: 
\begin{itemize}[leftmargin=*]
    \item We introduce a behaviorally grounded modeling paradigm that distinguishes bull and bear forces, empirically validating how market biases evolve and impact price trends across time and industries.
    \item Our model integrates time-series price data and external textual information by introducing the \emph{Bias-Aware Representation from Prices to Concept} and capture the interactions between Bullish and Bearish forces by \emph{Momentum-Aware Competitive Modeling}. The model generates more accurate and actionable trading insights based on real-world market dynamics. 
    \item Experimental results based on real-world financial data (comprising 110 trades across 10 industries) demonstrate that \ours\ excels in market trend prediction while revealing the complex interactions between investor biases and market behaviors. 
    \item Our work proposed a detailed case to demonstrate how evolving investor bias and attention—quantified through novel metrics—drive sentiment shifts, emotional polarization, and structural movements in financial markets.
\end{itemize}

The organization of our paper is as follows: 
we first present related work in Section \ref{sec:Relatedwork}.
Section \ref{sec:model} details how to effectively fuse voice interactions and knowledge graph information to mine user preferences.
Experimental results are provided in Section~\ref{sec:Experiments}.
For better understanding, detailed case analyses illustrating bias evolution and its market impact are provided in Section~\ref{sec:discussion}.
Finally, conclusions and future work are presented in Section \ref{sec:Conclusion}.

%% file: B-2-Related.tex
\section{Related Work}
\label{sec:Relatedwork}
\subsection{Multimodal Fusion Methods}
Financial forecasting has shifted from using only historical price data to integrating multimodal information such as news, social media, and reports. Early methods employed simple feature concatenation—\emph{e.g.}, combining prices with bag-of-words or sentiment scores \cite{xu2018stock}—which improved accuracy but overlooked temporal and semantic misalignments. 
Later, models like FinReport \cite{li2024FinReportES} and MSMF \cite{qin2024msmf} addressed alignment more explicitly, either by structuring textual events or dynamically adjusting modality weights. Recent efforts—such as CMTF \cite{pei2025crossmodaltemporalfusionfinancial}, MSGCA \cite{zong2024stockmovementpredictionmultimodal}, and modality-aware transformers \cite{emami2024modality}—further introduced cross-modal attention and semantic fusion, often leveraging pretrained models like FinBERT \cite{huang2023finbert} to align textual sentiment with numerical features \cite{mou2025mm}. 
Despite these advances, most attention mechanisms remain scalar and uninterpretable, failing to reflect opposing market views such as bullish versus bearish interpretations.

Graph-based and LLM-enhanced approaches extend fusion by modeling broader relational structures and narrative flows. 
Temporal GNNs and hierarchical graph models \cite{feng2019temporal, iyer2023bilevelattentiongraphneural, xiang2022temporal} capture inter-stock dependencies but typically collapse sentiment diversity into aggregated node states. 
Similarly, LLM-based frameworks like TIME-LLM \cite{jin2023time} improve alignment and generalization but encode market discourse as a single narrative stream, overlooking polarity and investor competition. Overall, while existing methods excel at collecting and aligning multimodal inputs, they rarely model the dynamics of opposing investor behavior—missing a critical link between sentiment divergence and price formation that our work aims to address.

\subsection{Market-Modeling Paradigms}
Price reflects the aggregated outcome of investor behavior, and its modeling traditionally relies on stochastic frameworks such as the random walk \cite{godfrey1964random}, GARCH models \cite{bollerslev1994arch}, and Black-Scholes option pricing \cite{black1973pricing}. 
Though effective in capturing volatility and short-term fluctuations, these models assume market efficiency and often neglect external information and investor heterogeneity, limiting their explanatory capacity during structural or sentiment-driven changes.

Extensions incorporating macro shocks (\emph{e.g.}, SVAR \cite{lin2024extra}), sentiment-aware GARCH variants \cite{fang2023sentiment}, and denoising techniques \cite{gulay2025predictive} improve robustness, but still treat behavior implicitly or linearly. Deep learning models—including LSTMs, Transformer hybrids \cite{zeng2023financial}, and Market-GAN \cite{xia2024market} better capture nonlinear dynamics, yet often model price as a closed system, leaving decision logic unaddressed.
Recent studies attempt to uncover behavioral signals: MF-CLR \cite{DuanZDWJQ24} uses contrastive learning for directional trends, AutoTimes \cite{liu2024autotimes} encodes candlestick patterns as symbolic sequences, and CMLF \cite{hou2021stock} models momentum via memory-aligned representations. However, few works explicitly separate bullish and bearish interpretations or explain how competing viewpoints drive market movement.
This leaves open the need for a model that extracts interpretable bias from external signals and links it to behavioral dynamics that shape price trajectories—a direction our \ours{} model pursues.

%% file: B-4-Model.tex
\section{Methodology}
\label{sec:model}
\begin{figure*}[htbp]
\centering
    \includegraphics[width=1\linewidth]{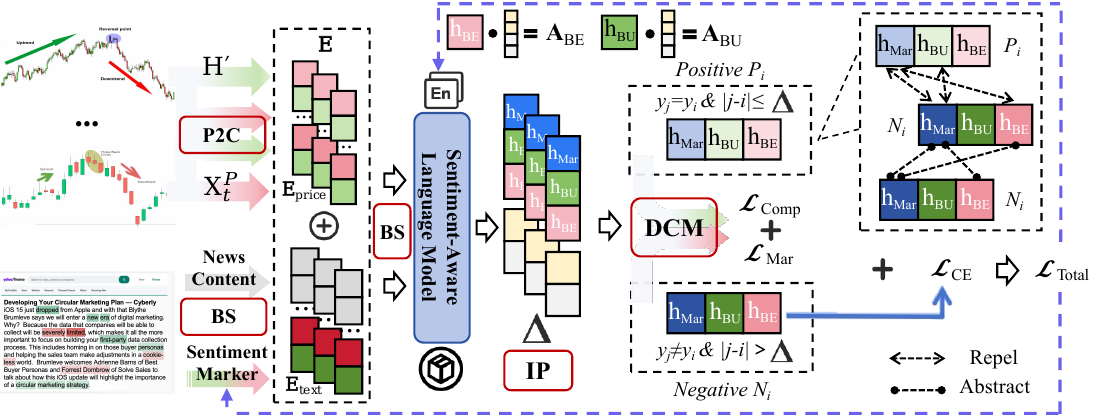}
    \caption{The overall architecture of \ours~model for dynamic market modeling. The framework is designed to address key questions regarding the integration of temporal data and the mitigation of bias to extract insights from news events, as well as capturing price momentum and reflecting bull-bear competition to produce actionable trading insights. 
    }
    \label{figs:framework}
\end{figure*}

To address the core modeling challenge of transforming bias-aware representations into behavior-oriented predictions, this section aims to bridge the gap between investors’ interpretations of market information and the resulting patterns of observable trading behavior.
To this end, we focus on two key questions:
\textbf{Q1}: How can temporal price signals be integrated with noisy external information sources to construct bias-aware representations?
\textbf{Q2}: How can market momentum and investor competition be effectively captured to simulate behavioral dynamics and generate actionable insights?

To address these questions, our framework (Figure~\ref{figs:framework}) comprises two coordinated modules: Bias-Aware Representation, which transforms price data into conceptual embeddings, and Momentum-Aware Competitive Modeling, which captures behavioral dynamics through competitive interactions.

\subsection{Bias-Aware Representation from Prices to Concept}

To effectively capture the interaction between price dynamics and external information sources (\emph{e.g.}, news, policy signals, and social media sentiment), we introduce two complementary modules: \emph{Price-to-Concept} (P2C) and \emph{Bias Simulation} (BS).
P2C aligns structured price time series with semantic space representations, enabling cross-modal integration, while BS captures investors’ divergent interpretations, bullish and bearish, of the same information. 
Together, they produce a bias-aware market representation that reflects both temporal volatility and subjective sentiment, forming the foundation for behavior-oriented financial modeling.


\par\noindent
\textbf{Price-to-Concept (P2C).}
Financial time series and textual news data belong to inherently different modalities. 
To align them, the P2C module first transforms historical price sequences into a latent representation compatible with the language model’s semantic space. 
The initial input is a multivariate price sequence $\mathbf{X}_{\mathit{t}}^{\mathit{P}}$ within a specified time window $[\mathit{t-\delta, t}]$, where $\mathit{\delta}$ denotes the length of the historical look-back window.
It includes the open price $\mathit{x}_{\mathit{t}}^{\text{Open}}$, close price $\mathit{x}_{\mathit{t}}^{\text{Close}}$, lowest price $\mathit{x}_{\mathit{t}}^{\text{Low}}$, and highest price $\mathit{x}_{\mathit{t}}^{\text{High}}$, which reflect historical trends and real-time market fluctuations. 
To encode price series into a conceptualized representation, we first extract the initial-layer output $\mathbf{H_t}$ from an Encoder-Only Language Model (EOLM), such as BERT or RoBERTa. 
Since time-series and text belong to different modalities with varying semantic expression methods, and Language Models cannot directly process time-series input features, it is necessary to align price sequence with the natural language text domain.

Given the vast space of the natural language text domain, we apply a \text{Linear} transformation to obtain the text prototype matrix $\mathbf{H'}$, which represents market-related semantic features: $\mathbf{H'} = \text{Linear}(\mathbf{H})$.
Next, the price window $\mathbf{X}_\mathit{t}^\mathit{{P}}$ interacts with $\mathbf{H'}$ through P2C to generate the semantic-related price embedding:
\begin{align}
    \label{P2C}
    \mathbf{E}_{\text{price}}
=\text{Softmax}\left(\frac{\mathbf{X}_\mathit{t}^\mathit{{P}} \cdot \mathbf{H'}^\top}{\sqrt{\mathit{d_k}}}\right) \cdot \mathbf{H^{'}},
\end{align}
where $\mathit{d_{k}}$ represents the dimension of linearly transformed vectors for scaled dot-product attention. This process aligns price fluctuations with textual prototypes, narrowing the gap between numerical market data and semantic understanding.

\par\noindent
\textbf{Bias Simulation (BS).}
While P2C enables semantic grounding of price sequences, the BS module captures asymmetric interpretations arising from distinct bullish and bearish perspectives of investors.

(1) \textbf{Sentiment Marker Injection:}
To simulate investor perspectives, we incorporate symbolic sentiment cues into news texts. Let $\mathbf{X}^{\mathit{N}}_\mathit{t}$ denote the external news on day $\mathit{t}$. We prepend two sentiment markers, `UP' and `DOWN', to the original news input to signal bullish and bearish viewpoints. 
Text enriched with sentiment markers is processed separately from price data. The augmented text data for day $\mathit{t}$, $\mathbf{X}_{\mathit{t}}^{\text{Aug}}$, is generated by concatenating the sentiment markers with the original text:
\begin{align}
\label{aug}
    \mathbf{X}^{\text{Aug}}_{\mathit{t}}=\mathrm{Concat}(\text{[UP]},\text{[DOWN]},\mathbf{X}^{N}_{\mathit{t}}).
\end{align}

The augmented text undergoes tokenization and embedding, which converts the text into a sequential representation: 
\begin{align}
    \mathbf{E}_{\text{text}} = \mathrm{f}(\mathbf{X}_{\mathit{t}}^{\text{Aug}}),
\end{align}
where $\mathrm{f}(\cdot)$ denotes the embedding and encoding function. The embedding layer of a pretrained EOLM, which in this case is the BERT embedding layer, maps the tokenized sequence into a high-dimensional embedding space $\mathbf{E}_{\text{text}}$. Unlike price data, textual sequences do not participate in direct cross-attention but act as complementary sentiment representations.

(2) \textbf{Bias Modeling and Representation:}
Investor biases, the directional divergence arising when bullish and bearish camps interpret the same signal differently, are embedded by fusing numerical prices with sentiment-injected text.

We first concatenate the price-aligned embedding \(\mathbf{E}_{\text{price}}\) and news embedding \(\mathbf{E}_\text{text}\) to obtain a joint feature matrix $\mathbf{E}$. 
\begin{align}
    \mathbf{E}=\text{Concat}(\mathbf{E}_{\text{text}}, \mathbf{E}_{\text{price}}).
\end{align}

To reflect the investors' varied understandings of the information, this unified representation $\mathbf{E}$ is passed through a Sentiment-Aware Language Model ($\mathcal{SAL}$), a pretrained EOLM, which has been shown effective for sentiment analysis due to its self-attention mechanism and ability to capture both local and global context. 
The encoder is particularly suitable in this setting because it dynamically weighs the importance of each token, enabling the model to interpret sentiment-bearing cues (\emph{e.g.}, emotional verbs, company names) within the broader market narrative. We express this sentiment encoding process as:
\begin{align}
    \mathbf{h}=\mathcal{SAL}(\mathbf{E}).
\end{align}

The output of this encoder is a contextualized sequence of embeddings $\mathbf{h} \in \mathbb{R}^{|\mathbf{E}|\times d}$, from which we extract three key components:
\begin{equation}\label{bb_index}
    \small
    \mathbf{h}_{\text{Mar}}= \mathbf{h}[\text{idx}(\text{[CLS]})], \quad \mathbf{h}_{\text{Comp}}=\mathbf{h}[\text{idx}(\text{[UP, DOWN]})].
\end{equation}
Investor sentiment is modeled by extracting two embeddings from the final layer output. 
$ \mathbf{h}_{\text{Mar}}$ represents the market representation, indexed at the $\text{[CLS]}$ token, while  $\mathbf{h}_{\text{Comp}}$ represents the competition representation which includes both bullish and bearish sentiments features, indexed at the $\text{[UP]}$ and $\text{[DOWN]}$ tokens. 
The sentiment markers $\text{[UP]}$ and $\text{[DOWN]}$ correspond to bullish and bearish sentiments, respectively, which represent opposing positions in the competition between market participants, driving the dynamics of price movements.

The indexing function $\text{idx}$ extracts the specific position of market and sentiment representations from $\mathbf{h}$. 
These sentiment-specific vectors, $\mathbf{h}_{\text{BU}}$ and $\mathbf{h}_{\text{BE}}$, are extracted from [\text{UP}] and [\text{DOWN}] fields of $\mathbf{h}_{\text{Comp}}$. 
These vectors are then combined with the overall representation $\mathbf{h}$, generating bias-aware representations:
\begin{equation}\label{attention}
    \mathbf{A}_{\text{BU}} = \frac{\mathbf{h}_{\text{BU}}\cdot   \mathbf{h}^\top}{\sqrt{\mathit{d}}} , \quad \mathbf{A}_{\text{BE}} = \frac{\mathbf{h}_{\text{BE}}\cdot  \mathbf{h}^\top}{\sqrt{\mathit{d}}}.
\end{equation}
where, the resulting biases can be formulated as the difference between these two attention maps:
$\mathbf{A}_{\text{BU}} - \mathbf{A}_{\text{BE}}$.

This bias matrix reflects how investor sentiment skews attention toward different features of the input, offering an interpretable and differentiable mechanism to simulate investor heterogeneity. 
It enables the model to account for the asymmetric impact of news and market signals on price actions, ultimately improving the understanding and prediction of dynamic market behavior.

\subsection{Momentum-Aware Competitive Modeling}
\label{sec:inertial}
Market momentum shows up as persistent runs in price direction, while bullish and bearish investors compete to extend or reverse those runs.  

To model both phenomena in a single pipeline, we propose an integrated strategy combining \emph{Inertial Pairing} (IP) and \emph{Dual Competition Mechanism} (DCM), capturing the competitive dynamics between bullish and bearish market trends while preserving the temporal consistency of sequential data. 
This strategy leverages temporal momentum to construct meaningful samples and employs contrastive losses to align trend-specific features with overall market characteristics.

\par\noindent
\textbf{Inertial Pairing (IP).} 
To capture market momentum, we introduce the concept of \emph{Inertial Pairing} to generate positive and negative samples. 
Positive samples maintain the same market trend over time, while negative samples either deviate from the trend or fall outside a given temporal window. 
This approach ensures that the model learns to account for the momentum inertia in financial markets, reflecting the persistence of price movements in the same direction.

(1) \textbf{Market Trend Labeling}: At each time $\mathit{t}$, we derive the true market trend label $\mathit{y_t}$ based on the direction of price movement, distinguishing between bullish and bearish trends. Specifically, we define:
\begin{equation}
    \label{mtl}
   \mathit{y_t} =
   \begin{cases}
   1, & \text{if } \mathit{x}_{\mathit{t}+1}^\text{Close} > \mathit{x}_\mathit{t}^{\text{Close}}, \\
   
   0, & \text{if } \mathit{x}_{\mathit{t}+1}^{\text{Close}} < \mathit{x}_\mathit{t}^{\text{Close}},
   \end{cases}
\end{equation}
where, $\mathit{y_t} = 1$ indicates a bullish trend, while $\mathit{y_t} = 0$ represents a bearish trend. This labeling is used to create positive and negative samples aligned with the market momentum.

(2) \textbf{Sample Momentum}: Momentum refers to the persistence of price movements in the same direction. 
In our model, we leverage market trend inertia to construct samples that capture this momentum, ensuring that the model can learn the temporal dependencies inherent in financial data. 

Positive samples $\mathcal{P}_i$ align with the current sample’s trend and fall within a predefined temporal range $\Delta$, while negative samples $\mathcal{N}_i$ either deviate from the trend or lie outside the temporal boundary:
\begin{align}
   \mathcal{P}_i &= \{ \mathit{x_{j}} \,|\, \mathit{y_j} = \mathit{y_i} \text{ and } |\mathit{j} - \mathit{i}| \leq \Delta \},
   \label{eq:Pi} \\
   \mathcal{N}_i &= \{ \mathit{x_{j}} \,|\, \mathit{y_j} \neq \mathit{y_i} \text{ or } |\mathit{j} - \mathit{i}| > \Delta \},
   \label{eq:Pn}
\end{align}
where the temporal range $\Delta$ encapsulates the market's momentum inertia, ensuring that positive and negative samples reflect sequential consistency.

\par\noindent
\textbf{Dual Competition Mechanism (DCM).} 
To model dynamic competition and its impact on market trends, we design a mechanism to establish dual object links, the competing participants (bullish vs. bearish) and their resulting outcomes (market direction). 
The motivation stems from the inherent interplay between bullish and bearish forces, which collectively drive the market toward bull or bear markets. 
Consequently, stronger bullish pressure is typically associated with upward price movements, whereas intensified bearish sentiment often precedes market downturns.

Given $N$ samples constructed by \emph{Inertial Pairing}, the \emph{Dual Competition Mechanism} employs contrastive losses for feature alignment to align bullish and bearish features with overall market dynamics. 
$\mathbf{h}_{\text{Comp},\mathit{i}}$ captures the competition between bullish and bearish forces, while $\mathbf{h}_{\text{Comp},\mathit{p}}$ reflects the dominance of bullish forces over bearish ones. $\mathbf{h}_{\text{Mar}}$, on the other hand, represents the market characteristics of bull or bear markets. 
The loss functions are defined as follows:
\begin{align}
    \label{dual}
    \centering
    \small
    \mathcal{L}_{\text{Comp}} &= \frac{1}{\mathit{N}} \sum_{i \in \mathcal{D}} \frac{1}{|\mathcal{P}_i|} \sum_{\mathit{p} \in \mathcal{P}_\mathit{i}} -\log \frac{\exp(\mathcal{H}_{CM}(p,i)/\tau)}{\sum_{\mathit{a} \in \mathcal{A}_{\mathit{i}}} \exp(\mathcal{H}_{CM}(a,i)/\tau)},
    \\
    \mathcal{L}_{\text{Mar}} &= \frac{1}{\mathit{N}} \sum_{\mathit{i} \in \mathcal{D}} \frac{1}{|\mathcal{P}_\mathit{i}|} \sum_{\mathit{p} \in \mathcal{P}_\mathit{i}} -\log \frac{\exp(\mathcal{H}_{CM}(i,p)/\tau)}{\sum_{\mathit{a} \in \mathcal{A}_\mathit{i}} \exp(\mathcal{H}_{CM}(i,a)/\tau)},
\end{align}
where $\mathcal{H}_{CM}(p,i)=\mathbf{h}_{\text{Comp},\mathit{p}}^\star\cdot  \mathbf{h}_{\text{Mar},\mathit{i}}$, 
$\mathcal{H}_{CM}(a,i)=\mathbf{h}_{\text{Comp},\mathit{a}}^\star \cdot \mathbf{h}_{\text{Mar},\mathit{i}}$, 
$\mathcal{H}_{CM}(i,p)=\mathbf{h}_{\text{Comp},\mathit{i}}^\star\cdot \mathbf{h}_{\text{Mar},\mathit{p}}$,
$\mathcal{H}_{CM}(i,a)=\mathbf{h}_{\text{Comp},\mathit{i}}^\star \cdot \mathbf{h}_{\text{Mar},\mathit{a}}$, 
$\tau$ is the temperature factor, 
$\mathcal{D}$ the set of indexes of the training samples, and $\mathcal{A}_i$ refers to the union of sets $\mathcal P_{i}$ and $N_{i}$. $\mathbf{h}_{\text{Comp},i}^\star$ denotes the columns of $\mathbf{h}_{\text{Comp},i}$ corresponding to the ground-truth label of $\mathbf{h}_{\text{Mar},i}$.

To quantify the momentum-aware dominance of bullish or bearish forces, a cross-entropy loss is introduced:
\begin{align}
    \label{ce}
    \small
    \mathcal{L}_{\text{CE}} = -\frac{1}{N} \sum_{i \in \mathcal{D}} \log \frac{\exp(\mathcal{H}_{CM}(i,i))}{\exp(\mathcal{H}_{UM}(i,i))+ \exp(\mathcal{H}_{EM}(i,i))},
\end{align}
where $\mathcal{H}_{CM}(i,i)=\mathbf{h}^\star_{\text{Comp},\mathit{i}} \cdot \mathbf{h}_{\text{Mar},\mathit{i}}$,
$\mathcal{H}_{UM}(i,i)=\mathbf{h}_{\text{BU},\mathit{i}} \cdot \mathbf{h}_{\text{Mar},\mathit{i}}$,
$\mathcal{H}_{EM}(i,i)=\mathbf{h}_{\text{BE},\mathit{i}} \cdot \mathbf{h}_{\text{Mar},\mathit{i}}$.


The final loss function combines the contrastive and momentum losses, balancing the objectives of trend alignment and feature competition:
\begin{align}
   \mathcal{L}_{\text{Total}} = \alpha \cdot (\mathcal{L}_{\text{Comp}} + \mathcal{L}_{\text{Mar}}) + (1 - \alpha) \cdot \mathcal{L}_{\text{CE}},
   \label{eq:um}
\end{align}
where $\alpha$ is the weighting parameter.

\par\noindent
\textbf{Model Optimization.}
The model parameters are optimized using backpropagation and gradient descent:
\begin{align}\label{optimize}
    \theta_{\text{new}} = \theta_{\text{old}} - \eta \nabla_{\theta} \mathcal{L}_{\text{Total}},
\end{align}
where $\eta$ is the learning rate. 
During training, the model minimizes $\mathcal{L}_{\text{Total}}$, ensuring not only accurate market trend predictions but also effective differentiation between bullish and bearish dynamics.

By combining \emph{Inertial Pairing} and \emph{Dual Competition Mechanism}, the framework establishes temporal consistency in sample construction and enhances the competitive relationship modeling between bullish and bearish forces. 
This integration allows the model to better capture market momentum and investor sentiment, reflecting the nuanced dynamics of financial markets.

%% file: B-5-1-Experiment.tex
\section{Experimental Evaluation}
\label{sec:Experiments}

In this section, we evaluate the performance of the \ours.
Extensive experiments carried out on 10 industries are designed to explore and answer the following research questions (\textbf{RQs}):
\begin{itemize}[leftmargin=*]
    \item \textbf{RQ1}: How does \ours~perform compare to other models using text and time series for stock price prediction?
    \item \textbf{RQ2}: Do the price translator and contrastive modules improve \ours~performance, and how do different contrastive losses affect it?
    \item \textbf{RQ3}: How do hyperparameter combinations impact \ours's performance, and which is optimal?
\end{itemize}

\input{tables/compare-y-dataset-metric-x-method}

\begin{figure*}[!t]

	\centering
        \begin{subfigure}[t]{0.9\linewidth}
		\includegraphics[width=\textwidth]{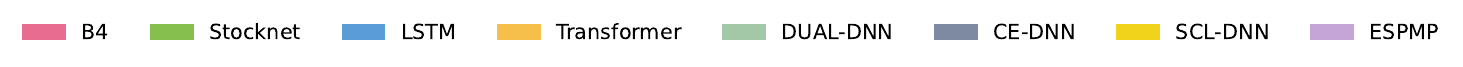}
		\label{fig:clr_ml_light}
	\end{subfigure}
	\begin{subfigure}[t]{0.24\linewidth}
		\includegraphics[width=\textwidth]{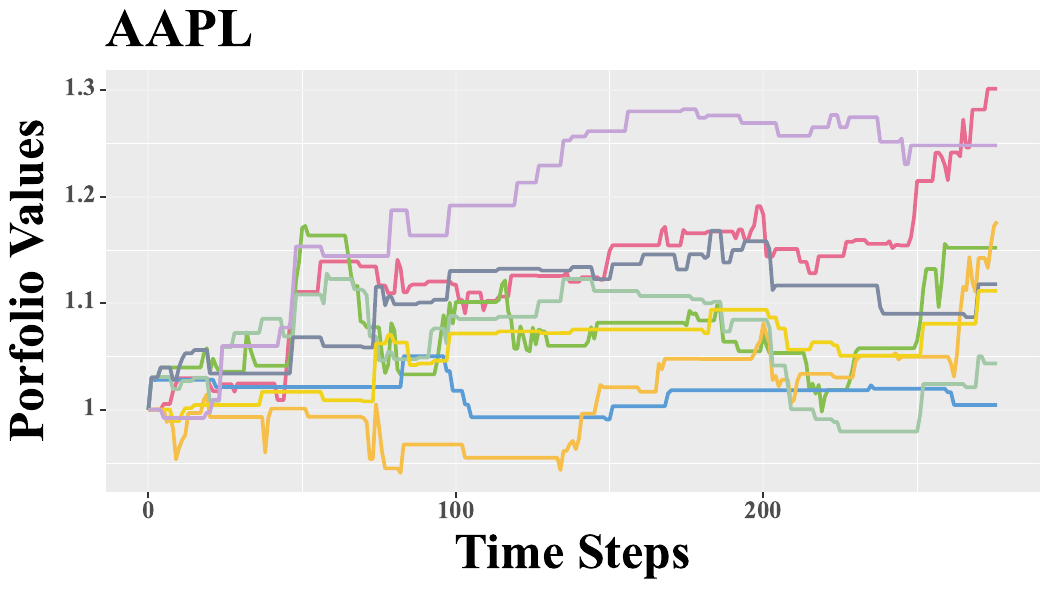}
		\label{fig:clr_ml_light}
	\end{subfigure}
	\begin{subfigure}[t]{0.24\linewidth}
		\includegraphics[width=\textwidth]{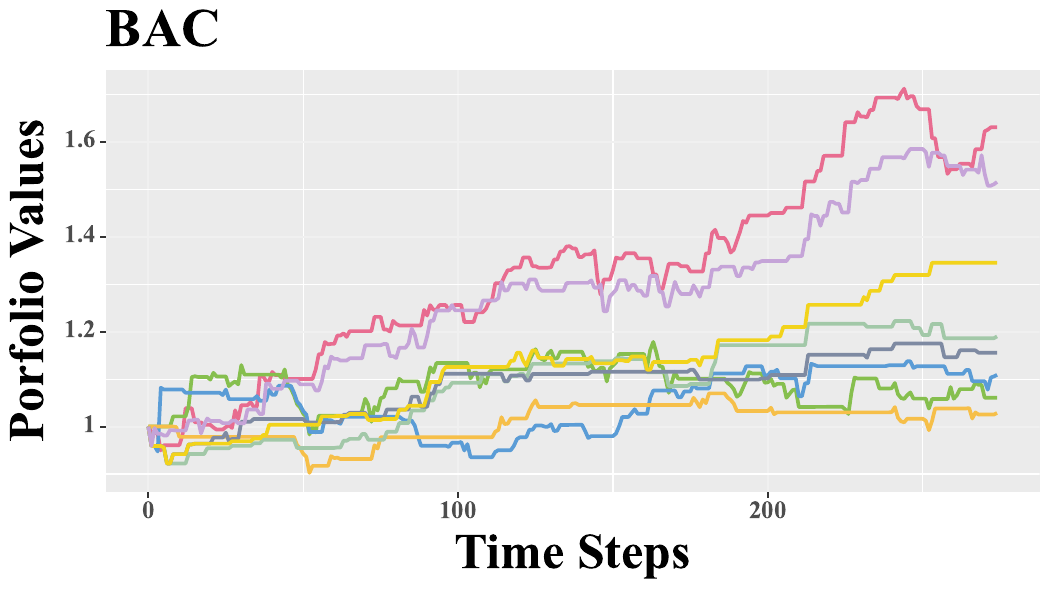}
		\label{fig:clr_ml_wmf}
	\end{subfigure}
	\begin{subfigure}[t]{0.24\linewidth}
		\includegraphics[width=\textwidth]{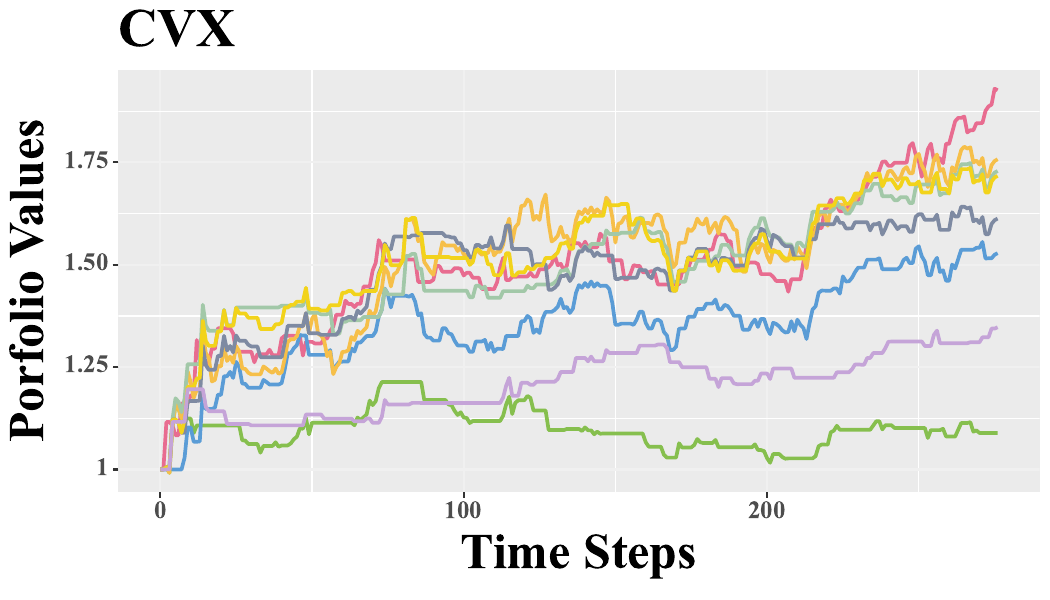}
		\label{fig:clr_yelp_light}
	\end{subfigure}
	\begin{subfigure}[t]{0.24\linewidth}
		\includegraphics[width=\textwidth]{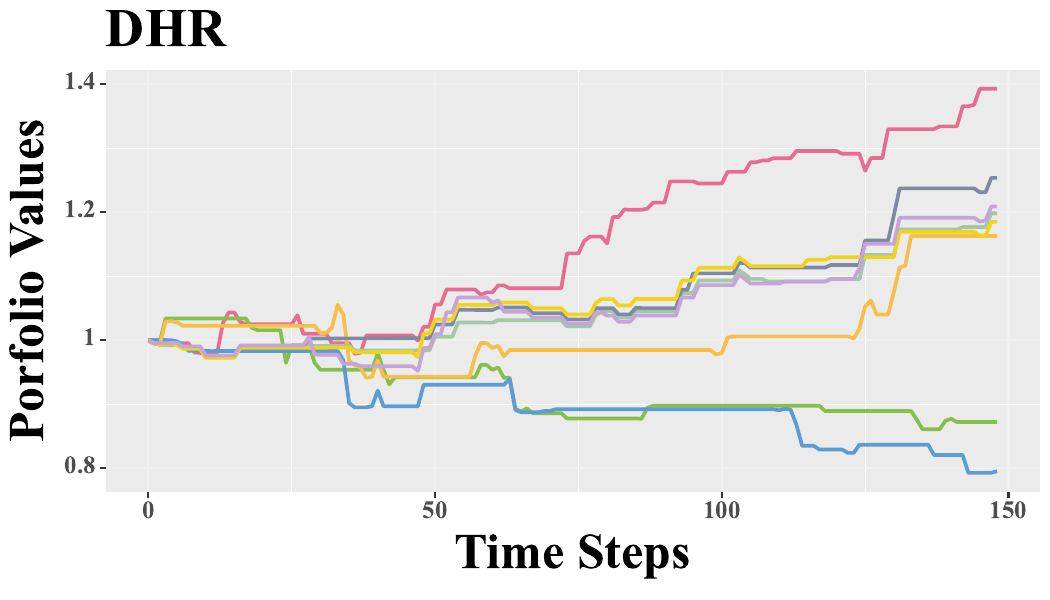}
		\label{fig:clr_yelp_wmf}
	\end{subfigure}
        \begin{subfigure}[t]{0.24\linewidth}
		\includegraphics[width=\textwidth]{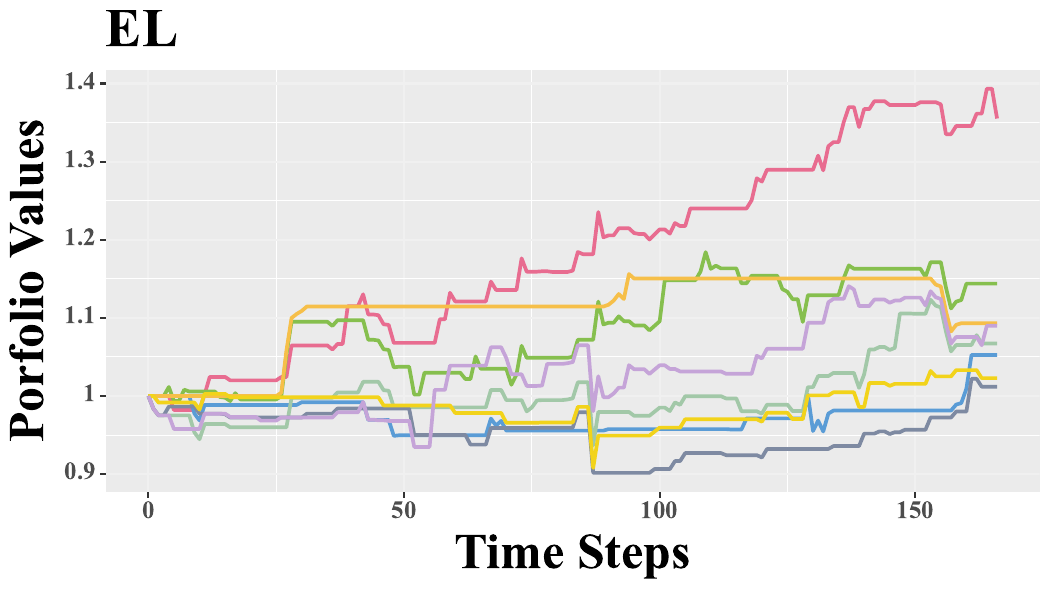}
		\label{fig:clr_ml_light}
	\end{subfigure}
	\begin{subfigure}[t]{0.24\linewidth}
		\includegraphics[width=\textwidth]{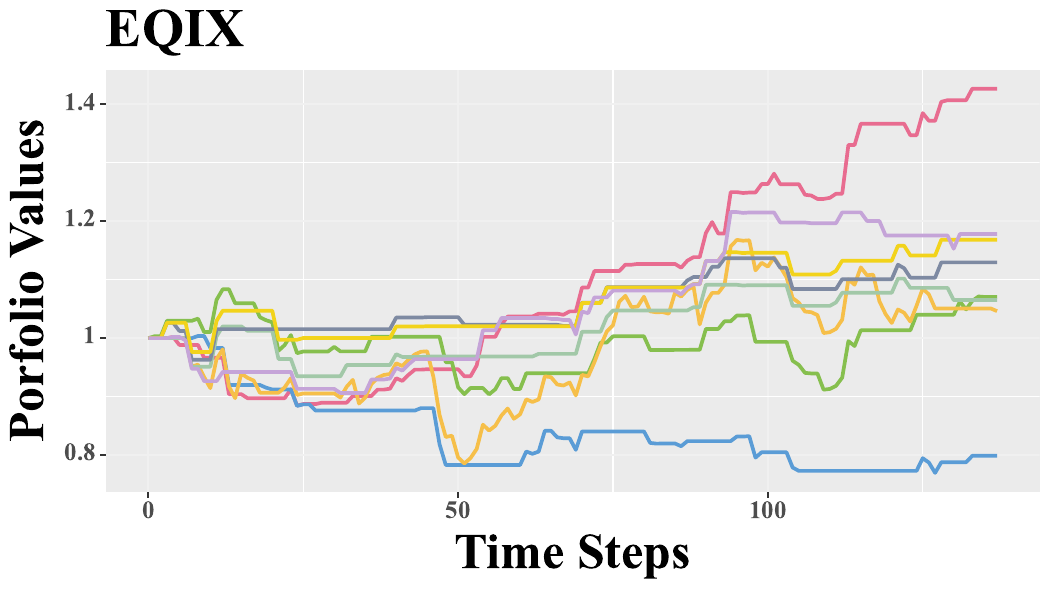}
		\label{fig:clr_ml_wmf}
	\end{subfigure}
	\begin{subfigure}[t]{0.24\linewidth}
		\includegraphics[width=\textwidth]{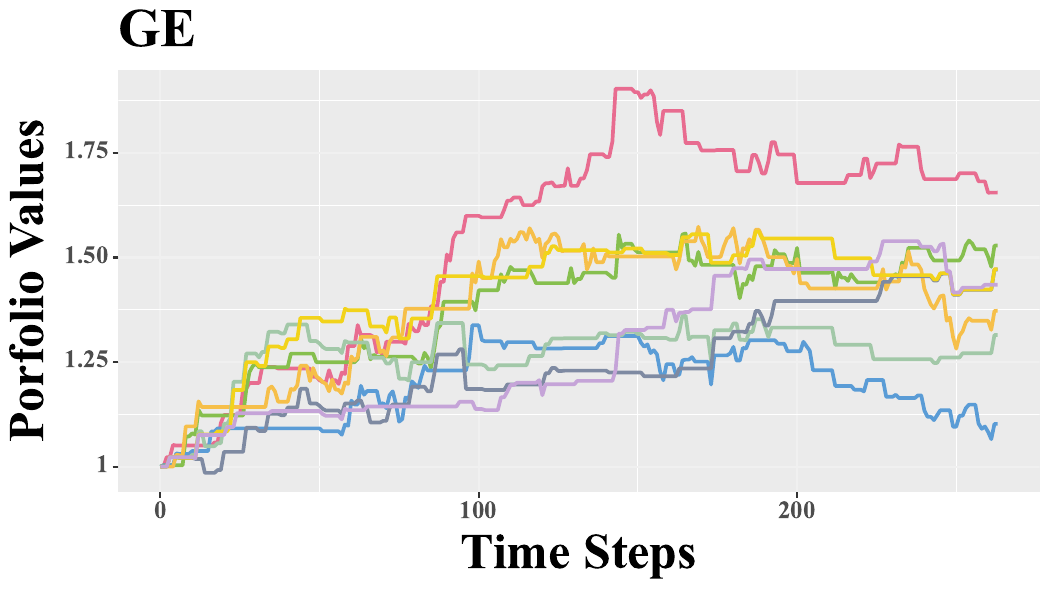}
		\label{fig:clr_yelp_light}
	\end{subfigure}
	\begin{subfigure}[t]{0.24\linewidth}
		\includegraphics[width=\textwidth]{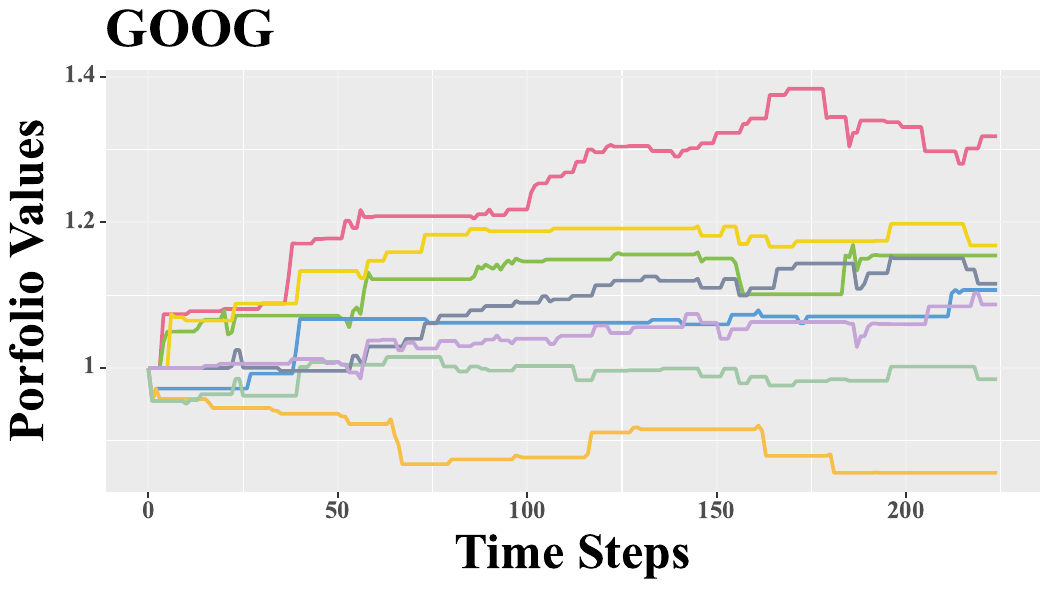}
		\label{fig:clr_yelp_wmf}
	\end{subfigure}
	\caption{
         Performance Comparison of magnitude $\eta$ of CL.
         \label{fig:clr}
	}
\end{figure*}

\subsection{Experiment Settings}
\par\noindent
\textbf{Datasets}.
CMIN dataset\footnote{\url{https://github.com/BigRoddy/CMIN-Dataset/}} comprises the top 110 stocks by market capitalization from the U.S. stock market datasets \cite{luo-etal-2023-causality}, categorized into 10 industries from 2018-01-01 to 2021-12-31. The industries covered include: 
1) Communication, 
2) Consumer, 
3) Energy, 
4) Financial, 
5) Healthcare, 
6) Industrials, 
7) Materials, 
8) Real Estate, 
9) Technology,
and 
10) Utilities. 
The correspondence between industries and stocks, are detailed in Table \ref{apx:industry_encoder}. The training set consists of the first 70\% of the time span for all stocks, while the testing set comprises the remaining 30\% of the time span.

\par\noindent
\textbf{Evaluation Metrics}.
We evaluate the risk and return characteristics of the trading strategy from multiple perspectives by employing a set of performance metrics, including 
\emph{Annual Return}, \emph{Cumulative Returns}, \emph{Max Drawdown}, and \emph{Calmar Ratio}. 

In this study, we employ a set of performance metrics to assess the performance of our trading strategy. Each metric offers a unique perspective on the risk and return characteristics of the strategy. Below, we define each metric and provide a clear explanation of the variables involved in their respective formulas.

\begin{enumerate}[leftmargin=*]
    \item 
    \textbf{Annual Return}: The annual return is calculated by taking the geometric average of the return over the period and subtracting one. 
    It represents the average annual profit or loss from an investment:
    \begin{equation}
        \text{Annual Return} = \left(\frac{R}{n}\right)^{\frac{1}{t}} - 1, \nonumber
    \end{equation}
    where $R$ is the total return over the period, $n$ is the number of sub-periods (
    \textit{e.g.}, years), and $t$ is the number of years.

    \item \textbf{Cumulative Returns}: Cumulative returns measure the total change in the value of an investment over a specific period. 
    It is the ratio of the final value to the initial value minus one:
    \begin{equation}
        \text{Cumulative Returns} = \frac{V_t}{V_0} - 1,
        \nonumber
    \end{equation}
    where $ V_t $ is the final value of the investment, and $ V_0 $ is the initial value.

    \item \textbf{Max Drawdown}: 
    The maximum drawdown is the largest peak-to-trough decline in the value of a portfolio during a specific period before a new peak is achieved:
    \begin{equation}
        \text{Max Drawdown} = \max\left(\frac{P_{\text{peak}}}{P_{\text{valley}}} - 1\right),
        \nonumber
    \end{equation}
    where $P_{\text{peak}}$ is the highest peak value, and $P_{\text{valley}}$ is the lowest trough value following that peak.

    \item \textbf{Calmar Ratio}: 
    The Calmar ratio is a measure of risk-adjusted return that uses the maximum drawdown to represent downside risk:
    \begin{equation}
        \text{Calmar Ratio} = \frac{\text{Annual Return}}{\text{Max Drawdown}}.
        \nonumber
    \end{equation}
\end{enumerate}

These metrics are essential for providing a balanced view of the trading strategy's performance, considering both its profitability and its risk characteristics.

\par\noindent
\textbf{C. Benchmarks}.
We compare \ours~with seven representative baseline models, which fall into two major categories: 
1) Modal Fusion methods \textbf{(MF)}, which combine textual and price modalities for prediction, and 
2) Contrastive Learning-based Language models \textbf{(CL)}, which leverage contrastive objectives to model the competitive nature of market dynamics.

\begin{itemize}[leftmargin=*]
    \item \textbf{StockNet}~(MF)~\cite{xu2018stock}: A bidirectional Gated Recurrent Unit (GRU) to extract tweet information and aggregates price and text using an attention mechanism for prediction.
    \item \textbf{LSTM}~(MF)~\cite{lstm,zou2024novel}: A time series prediction model that directly concatenates text and price data.
    \item \textbf{Transformer}~(MF)~\cite{vaswani2023attentionneed,gao2023stockformer}: A prediction method utilizing attention mechanisms to capture global dependencies.
    \item \textbf{ESPMP}~(CL)~\cite{Xing2024ExplainableSP}: A novel triplet network for contrastive learning to enhance the explainability of stock movement prediction.
    \item \textbf{SCLformer}~(CL)~\cite{khosla2021supervisedcontrastivelearning}: A algorithm trading method based on supervised
    contrastive learning for sentiment recognition in text.
    \item \textbf{DUALformer}~(CL)~\cite{chen2022dual}: A trading model that uses a triple loss function for feature representation and trading decisions.
    \item \textbf{CEformer}~(CL)~\cite{8237586,Yang2024ALM}): A trading model that utilizes the cross-entropy loss function for binary classification tasks on news text.
\end{itemize}

\par\noindent
\textbf{D. Implementation Details}.
We account for the differences across various stock text datasets by using a fine-tuning approach to optimize the encoding performance for each dataset. 
We employed a BERT-based model with an uncased tokenization scheme to encode textual data associated with each company's stock. A distinct model was trained for each stock, with each model undergoing 200 training epochs to ensure convergence. For each stock, a dedicated model was instantiated and trained. 
We conducted a grid search to optimize the hyperparameters crucial for performance, which are $\alpha$ in Eq. \eqref{eq:um} and \texttt{MomentumSpan} $\Delta$ in Eqs. \eqref{eq:Pi} and \eqref{eq:Pn}. 
\begin{itemize}[leftmargin=*]
    \item $\alpha \in \{0.1, 0.3, 0.5, 0.7, 0.9\}$ balances two loss components: the contrastive alignment loss ($L_{\text{S}-\text{M}} + L_{\text{M}-\text{S}}$), which encourages consistency between sentiment and market features, and the cross-entropy loss ($L_{\text{CE}}$), which enhances polarity discrimination. A higher $\alpha$ emphasizes feature alignment, while a lower $\alpha$ focuses on distinguishing bullish and bearish dominance.
    \item $\Delta \in \{-2, -1, 0, 1, 2, \pm 1, \pm 2\}$ controls temporal sampling for market inertia: $\Delta = \pm 1$ to include immediate adjacent steps (past and future), $\Delta = -2$ to consider only two preceding steps, and $\Delta = \pm 2$ to span two steps before and after the current sample. This enables adaptive smoothing of market state continuity across heterogeneous entities.
\end{itemize}

The grid search strategy involved systematically combining different values of $\alpha$ and $\Delta$ to identify the optimal configuration that yielded the best performance on a validation set.

%% file: tables/compare-y-dataset-metric-x-method.tex
\begin{table*}[!t]
\centering
\caption{Performance comparison of different models across different stock datasets, with each stock representing a distinct industry. The table presents the Cumulative Returns, Annual Returns, Max Drawdown, and Calmar Ratio for each model, including \ours, LSTM, StockNet, Transformer, ESPMP, DUALformer, and SCLformer. We stylize the best result as \textbf{Bold} and the second best as \underline{underline}.}
\label{compare-each-stock}
\renewcommand{\arraystretch}{1.2}
\setlength{\tabcolsep}{6pt}
\scalebox{0.8}{
    \begin{tabular}{c|c|l|c|c|c|c|c|c|c|c} 
    \toprule
    \midrule
    \textbf{Industry} & \textbf{Stock} & \textbf{Metric} & \textbf{\ours} & \textbf{LSTM} & \textbf{Transformer} & \textbf{StockNet} & \textbf{DUALformer} & \textbf{CEformer} & \textbf{SCLformer} & \textbf{ESPMP} \\
    \midrule
    \multirow[t]{4}{*}{Technology} & \multirow[t]{4}{*}{AAPL} & Cumulative Returns \textuparrow& \textbf{0.301} & 0.005 & 0.177 & 0.152 & 0.043 & 0.118 & 0.112 & \underline{0.248} \\
     &  & Annual Return \textuparrow& \textbf{0.270} & 0.004 & 0.160 & 0.137 & 0.039 & 0.107 & 0.101 & \underline{0.223} \\
     &  & Max Drawdown \textdownarrow& 0.063 & 0.059 & 0.075 & 0.174 & 0.148 & 0.081 & \textbf{0.043} & \underline{0.052} \\
     &  & Calmar Ratio \textuparrow& \underline{4.290} & 0.072 & 2.133 & 0.788 & 0.266 & 1.319 & 2.342 & \textbf{4.338} \\
     \midrule
     \multirow[t]{4}{*}{Financial} & \multirow[t]{4}{*}{SCHW} & Cumulative Returns \textuparrow & \underline{0.697} & -0.007 & \textbf{1.108} & 0.270 & 0.018 & 0.232 & -0.006 & 0.158 \\
     &  & Annual Return \textuparrow & \underline{0.763} & -0.007 & \textbf{1.224} & 0.293 & 0.020 & 0.251 & -0.006 & 0.171 \\
     &  & Max Drawdown \textdownarrow & 0.123 & \textbf{0.041} & 0.240 & 0.176 & 0.104 & \underline{0.067} & 0.185 & 0.158 \\
     &  & Calmar Ratio \textuparrow & \textbf{6.223} & -0.180 & \underline{5.110} & 1.659 & 0.190 & 3.744 & -0.035 & 1.077 \\
     \midrule
     \multirow[t]{4}{*}{Energy} & \multirow[t]{4}{*}{CVX} & Cumulative Returns \textuparrow& \textbf{0.926} & 0.528 & \underline{0.758} & 0.089 & 0.729 & 0.613 & 0.717 & 0.347 \\
     &  & Annual Return \textuparrow& \textbf{0.812} & 0.471 & \underline{0.670} & 0.081 & 0.646 & 0.545 & 0.635 & 0.312 \\
     &  & Max Drawdown \textdownarrow & \underline{0.143} & 0.167 & 0.189 & 0.197 & 0.157 & 0.159 & 0.218 & \textbf{0.104} \\
     &  & Calmar Ratio \textuparrow& \textbf{5.685} & 2.818 & 3.547 & 0.411 & \underline{4.118} & 3.435 & 2.914 & 2.999 \\
    \midrule
    \multirow[t]{4}{*}{Real Estate} & \multirow[t]{4}{*}{EQIX} & Cumulative Returns \textuparrow & \textbf{0.427} & -0.201 & 0.046 & 0.071 & 0.065 & 0.130 & \underline{0.168} & 0.178 \\
     &  & Annual Return \textuparrow& \textbf{0.914} & -0.337 & 0.086 & 0.133 & 0.122 & 0.249 & 0.329 & \underline{0.349} \\
     &  & Max Drawdown \textdownarrow& 0.118 & 0.234 & 0.214 & 0.180 & 0.086 & \underline{0.063} & \textbf{0.050} & 0.096 \\
     &  & Calmar Ratio \textuparrow& \textbf{7.778} & -1.437 & 0.401 & 0.738 & 1.429 & 3.933 & \underline{6.567} & 3.634 \\
      \midrule
    \multirow[t]{4}{*}{Communication} & \multirow[t]{4}{*}{NFLX} & Cumulative Returns \textuparrow& \textbf{0.323} & 0.194 & 0.114 & \underline{0.319} & 0.129 & 0.046 & -0.003 & 0.213 \\
     &  & Annual Return \textuparrow& \underline{0.378} & 0.291 & 0.132 & \textbf{0.490} & 0.150 & 0.052 & -0.004 & 0.248 \\
     &  & Max Drawdown \textdownarrow& \textbf{0.073} & 0.132 & 0.192 & 0.375 & 0.119 & 0.100 & 0.093 & \underline{0.077} \\
     &  & Calmar Ratio \textuparrow& \textbf{5.169} & 2.195 & 0.689 & 1.306 & 1.261 & 0.525 & -0.042 & \underline{3.204} \\
       \midrule
    \multirow[t]{4}{*}{\centering Consumer} & \multirow[t]{4}{*}{\centering BABA} & Cumulative Returns \textuparrow & \textbf{0.119} & -0.089 & -0.066 & -0.162 & -0.167 & -0.281 & -0.304 & \underline{-0.043} \\
     &  & Annual Return \textuparrow & \textbf{0.126} & -0.098 & -0.069 & -0.170 & -0.175 & -0.293 & -0.317 & \underline{-0.046} \\
     &  & Max Drawdown \textdownarrow & \textbf{0.137} & 0.254 & 0.285 & 0.351 & 0.340 & 0.457 & 0.437 & \underline{0.137} \\
     &  & Calmar Ratio \textuparrow & \textbf{0.915} & -0.386 & \underline{-0.243} & -0.485 & -0.514 & -0.642 & -0.726 & -0.333 \\
       \midrule
    \multirow[t]{4}{*}{\centering Healthcare} & \multirow[t]{4}{*}{\centering PFE} & Cumulative Returns \textuparrow& \textbf{0.457} & -0.106 & 0.098 & -0.029 & -0.015 & 0.039 & 0.030 & \underline{0.262} \\
     &  & Annual Return \textuparrow& \textbf{0.502} & -0.114 & 0.106 & -0.031 & -0.016 & 0.042 & 0.032 & \underline{0.286} \\
     &  & Max Drawdown \textdownarrow & 0.095 & 0.135 & 0.232 & \textbf{0.029} & 0.115 & 0.126 & \underline{0.066} & 0.152 \\
     &  & Calmar Ratio \textuparrow& \textbf{5.273} & -0.842 & 0.459 & -1.080 & -0.137 & 0.333 & 0.488 & \underline{1.884} \\
       \midrule
    \multirow[t]{4}{*}{\centering Industrials} & \multirow[t]{4}{*}{\centering DE} & Cumulative Returns \textuparrow& \textbf{0.513} & 0.093 & 0.283 & -0.004 & 0.129 & 0.169 & \underline{0.308} & 0.252 \\
     &  & Annual Return \textuparrow& \textbf{0.680} & 0.119 & \underline{0.369} & -0.005 & 0.165 & 0.218 & 0.402 & 0.327 \\
     &  & Max Drawdown \textdownarrow & \underline{0.122} & 0.128 & 0.253 & 0.283 & 0.167 & \textbf{0.101} & 0.199 & 0.167 \\
     &  & Calmar Ratio \textuparrow& \textbf{5.573} & 0.928 & 1.462 & -0.019 & 0.987 & \underline{2.154} & 2.020 & 1.957 \\
       \midrule
    \multirow[t]{4}{*}{\centering Materials} & \multirow[t]{4}{*}{\centering BHP} & Cumulative Returns \textuparrow& \textbf{0.256} & 0.039 & 0.022 & -0.094 & \underline{0.140} & -0.048 & 0.088 & 0.083 \\
     &  & Annual Return \textuparrow& \textbf{0.342} & 0.052 & 0.029 & -0.120 & \underline{0.186} & -0.062 & 0.115 & 0.110 \\
     &  & Max Drawdown \textdownarrow & \textbf{0.057} & \underline{0.073} & 0.507 & 0.399 & 0.077 & 0.217 & 0.165 & 0.129 \\
     &  & Calmar Ratio \textuparrow & \textbf{6.017} & 0.702 & 0.058 & -0.300 & \underline{2.422} & -0.284 & 0.699 & 0.852 \\
    \midrule
    \multirow[t]{4}{*}{\centering Utilities} & \multirow[t]{4}{*}{\centering DUK} & Cumulative Returns \textuparrow& \textbf{0.370} & \underline{0.279} & 0.118 & 0.080 & 0.048 & -0.034 & 0.028 & 0.046 \\
     &  & Annual Return \textuparrow& \textbf{0.388} & \underline{0.294} & 0.124 & 0.084 & 0.050 & -0.036 & 0.030 & 0.048 \\
     &  & Max Drawdown \textdownarrow& \textbf{0.023} & 0.080 & \underline{0.035} & 0.077 & 0.055 & 0.096 & 0.066 & 0.056 \\
     &  & Calmar Ratio \textuparrow& \textbf{16.932} & \underline{3.668} & 3.531 & 1.081 & 0.917 & -0.373 & 0.453 & 0.861 \\
    \midrule
    \bottomrule
\end{tabular}
}
\end{table*}

%% file: B-5-2-Experiment-Analysis.tex
\subsection{Overall Preformance (RQ1)}
We provide examples from various industries to serve as intuitive illustrations of the advantages of our proposed algorithm. 
Table \ref{compare-each-stock} presents a performance comparison between the proposed \ours\ and other baseline methods. 
From the table, several key observations can be noted as follows:

\par\noindent
\textbf{Compare with TMF:} \ours~demonstrates competitive performance compared to traditional fusion methods like StockNet, LSTM, and Transformer. 
While these baseline models often excel in Cumulative Returns and Annual Return, \ours\ presents a more balanced performance across both returns and risk metrics. 
For instance, in the AAPL dataset, StockNet and LSTM achieve Cumulative Returns of 0.152 and 0.005, respectively, compared to \ours's 0.301. 
This suggests that \ours~outperforms these fusion methods in capturing long-term stock trends.
However, in terms of Max Drawdown and Calmar Ratio, traditional methods like LSTM and StockNet tend to show more substantial fluctuations, whereas \ours exhibits more stability, indicating its capability in controlling risk and optimizing the return-risk tradeoff.

\par\noindent
\textbf{Compare with SoTA:} 
\ours\ performs exceptionally well in comparison to contrastive learning-based methods, particularly in terms of Cumulative Returns and Annual Return. However, contrastive learning models like ESPMP, SCLformer, and DUALformer tend to show stronger performance in Max Drawdown and Calmar Ratio, which reflects their emphasis on learning risk-adjusted returns. 
In the EQIX dataset, CEformer achieves a Calmar Ratio of 4.597, significantly higher than \ours's 2.470, indicating that \ours~focus more on reducing risk while sacrificing some long-term profitability.

Overall, \ours\ strikes a balance between high returns and low risk. Although it shows slight shortcomings in certain risk-adjusted return metrics, it demonstrates robust and highly adaptive performance across multi-industry data scenarios.

\subsection{Ablation Analysis (RQ2)}
Four tasks are conducted to ascertain the significance and efficacy of the core components by removing \emph{Price-to-Concept} (w/o P2C), \emph{Bias Simulation} (\emph{w/o} BS), \emph{Inertial Pairing} (\emph{w/o} IP), and \emph{Dual Competition Mechanism} (\emph{w/o} DCM), respectively.

\input{pic_to_tex/ablation_barplot}

Figure \ref{figs:ablation_barplot} and Table \ref{ablation-each-stock} presents the overall performance of various model variants across different evaluation metrics.
The \ours~model (Full) consistently outperforms its ablated counterparts in both Cumulative Returns and Annual Returns matrix, highlighting the synergistic benefits of an integrated approach.
\ours~\emph{w/o} DCM exhibits the weakest performance, lacking the ability to discern market trends and thus failing to capitalize on the contrastive learning's potential.
\ours~\emph{w/o} BS shows improved performance, suggesting that \emph{Bias Simulation} significantly enhances the model's profitability.
\ours~\emph{w/o} IP performs better than models without \emph{Dual Competition Mechanism} or \emph{Bias Simulation}, indicating the module's role in effectively utilizing price information.
\ours~\emph{w/o} P2C performs secondary to the full model, indicating that while P2C contributes positively, it is the trend that most significantly impacts profitability.
\input{tables/ablation-y-dataset-metric-x-method}
For individual stock analysis, diving into individual datasets reveals variances that underscore the tailored impact of each module:
For AAPL, the full model achieves a Cumulative Return of 0.301, with \ours~\emph{w/o} BS lagging at -0.015, showcasing the critical role of BS. 
\emph{w/o} P2C manages a respectable 0.043, underlining the value of price information.
For NFLX, the full model excels with a Cumulative Return of 0.323, and \ours~\emph{w/o} DCM follows closely with 0.213, indicating that \ours~\emph{w/o} DCM is particularly influential for this stock.

\subsection{Hyper-parameters Combinations (RQ3)}
\input{pic_to_tex/bubble_pink_blue_plot} 
We further analyze the effects of the two key hyper-parameters introduced in Implementation Details: the loss-balancing weight $\alpha$ and the momentum window $\Delta$.
The goal is to reveal the interaction between $\alpha$ and $\Delta$ (Fig. \ref{figs:bubble}) as well as their sensitivity (Fig. \ref{figs:parameter_boxplot}). 

\emph{First}, when $\alpha$ is set to lower values (\textit{e.g.}, 0.1 and 0.3), the model's performance remains relatively stable across different values of $\Delta$, indicating that under low signal extraction weights, the model's dependence on momentum consistency is weak. 
However, as $\alpha$ increases to 0.7 or 0.9, the model's performance becomes significantly influenced by changes in $\Delta$, showing greater volatility. 
This suggests that under high signal extraction weights, the model's sensitivity to momentum span increases.
When $\Delta = \pm 2$ (\textit{i.e.}, a larger time-step range), the model outperforms the case where $\Delta = \pm 1$, demonstrating that an increased momentum span positively contributes to capturing market trends. 
In contrast, when $\Delta = 0$, the model's performance is moderate, suggesting that completely ignoring momentum consistency may weaken the model's ability to extract trend information.
The optimal performance is achieved under a combination of high $\alpha$ and $\Delta = \pm 2$, indicating that emphasizing the signal extraction task while incorporating a larger momentum span enhances model effectiveness. Additionally, competitive results are observed with a lower $\alpha$ and $\Delta = \pm 1$, suggesting that finding a balance between signal extraction and momentum consistency is also a viable strategy.

\input{pic_to_tex/parameter_each}

\emph{Second}, as shown in Figure \ref{figs:parameter_boxplot}, the sensitivity to hyperparameters varies across different stocks. 
In the AAPL dataset, both $\alpha$ and $\Delta$ have a significant impact on annualized returns. 
As $\alpha$ increases from 0.1 to 0.9, the median annualized return gradually improves, indicating that increasing the weight of the signal extraction task helps enhance the model's performance. 
The highest median annualized return is observed at $\Delta = \pm 1$, suggesting that moderate momentum consideration is most effective for the AAPL data.
In contrast, the ECL dataset exhibits relatively stable performance in annualized returns across different values of $\alpha$, with slight improvements only when $\alpha$ is set to intermediate values (\textit{e.g.}, 0.5 or 0.7). 
This suggests that ECL is less sensitive to the signal extraction task compared to AAPL. 
Additionally, the best performance for ECL occurs when $\Delta = 0$, indicating that for this stock, incorporating too much forward and backward time-step information may lead to overfitting or increased noise, resulting in a negative impact from momentum consistency.

%% file: pic_to_tex/ablation_barplot.tex

\begin{figure}[!t]
	\centering
	\begin{subfigure}[t]{0.48\linewidth}
		\includegraphics[width=\textwidth]{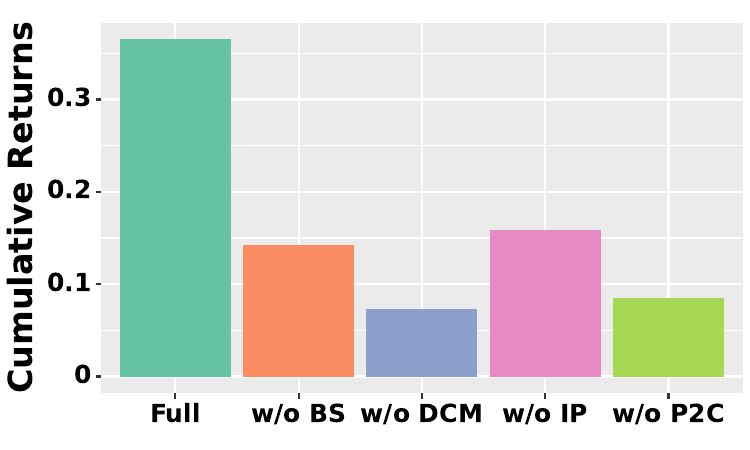}
		\label{fig:part_type_light}
	\end{subfigure}
	\begin{subfigure}[t]{0.48\linewidth}
		\includegraphics[width=\textwidth]{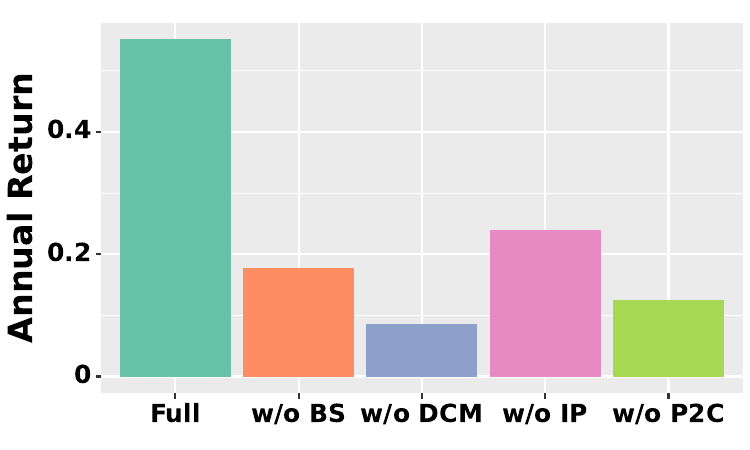}
		\label{fig:part_type_wmf}
	\end{subfigure}
    \begin{subfigure}[t]{0.48\linewidth}
		\includegraphics[width=\textwidth]{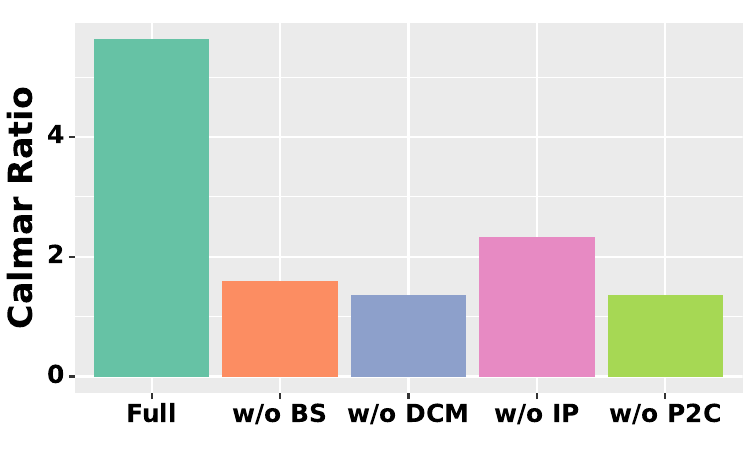}
		\label{fig:part_type_wmf}
	\end{subfigure}
    \begin{subfigure}[t]{0.48\linewidth}
		\includegraphics[width=\textwidth]{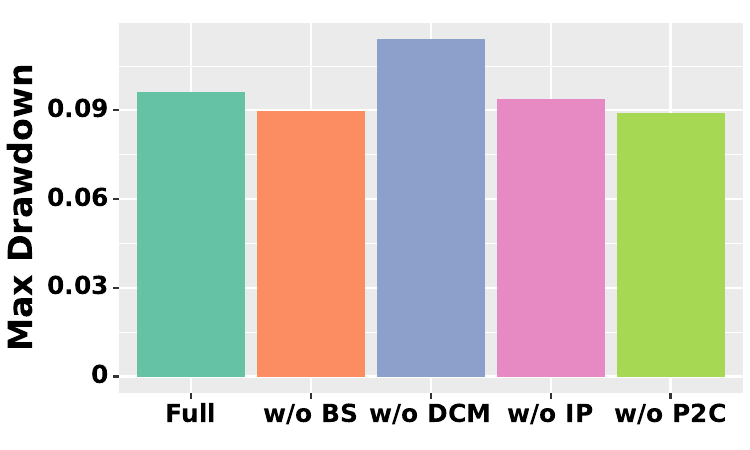}
		\label{fig:part_type_wmf}
	\end{subfigure}
	\caption{
       \textcolor{black}{Ablation study on key components of the proposed method, evaluating the impact of removing Price2Concept (P2C), Bias Simulation (BS), Inertial Pairing (IP), and Dual Competition Mechanism (DCM).}
    \label{figs:ablation_barplot}
	}
\end{figure}

%% file: tables/ablation-y-dataset-metric-x-method.tex
\begin{table}[!t]
\centering
\caption{Performance comparison on 5 model variants (Full, \emph{w/o} P2C, \emph{w/o} BS, \emph{w/o} IP, and \emph{w/o} DCM) across four datasets. We stylize the best result as \textbf{Bold} and the second best as \underline{underline}.}
\label{ablation-each-stock}
\setlength{\tabcolsep}{6pt}
\scalebox{.6}{
    \begin{tabular}{>{\centering\arraybackslash}m{1.1cm}|l|c|c|c|c|c}
    \toprule
    \midrule
    \textbf{Dataset} & \textbf{metric} &   \textbf{Full}   &   \textbf{\emph{w/o} P2C}   &   \textbf{\emph{w/o} BS}   &  \textbf{\emph{w/o} IP}    &  \textbf{\emph{w/o} DCM}   \\
    \midrule
    \multirow{4}{*}{AAPL} & Cumulative Returns \textuparrow & \textbf{0.301} & 0.043 & -0.015 & \underline{0.109} & 0.047 \\
       &  Annual Return \textuparrow & \textbf{0.270} & 0.039 & -0.013 & \underline{0.098} & 0.042 \\
       &  Max Drawdown \textdownarrow & 0.063 & 0.148 & \textbf{0.039} & 0.041 & \underline{0.039} \\
       &  Calmar Ratio \textuparrow & \textbf{4.290} & 0.266 & -0.346 & \underline{2.381} & 1.076 \\
    \midrule
    
    \multirow{4}{*}{SCHW} & Cumulative Returns \textuparrow & \underline{0.697} & 0.018 & \textbf{0.891} & 0.261 & 0.256 \\
       & Annual Return \textuparrow & \underline{0.763} & 0.020 & \textbf{0.980} & 0.283 & 0.277 \\
       & Max Drawdown \textdownarrow & \underline{0.123} & \textbf{0.104} & 0.233 & 0.149 & 0.156 \\
       & Calmar Ratio \textuparrow & \textbf{6.223} & 0.190 & \underline{4.203} & 1.899 & 1.778 \\
    \midrule
    
    \multirow{4}{*}{CVX} & Cumulative Returns \textuparrow & \textbf{0.926} & \underline{0.729} & 0.023 & 0.262 & -0.107 \\
    & Annual Return \textuparrow & \textbf{0.812} & \underline{0.646} & 0.020 & 0.235 & -0.098 \\
    & Max Drawdown \textdownarrow & \underline{0.143} & 0.157 & 0.176 & \textbf{0.075} & 0.193 \\
    & Calmar Ratio \textuparrow & \textbf{5.685} & \underline{4.118} & 0.116 & 3.123 & -0.507 \\
    \midrule
    
    \multirow{4}{*}{EQIX} & Cumulative Returns \textuparrow & \textbf{0.427} & 0.065 & -0.123 & \underline{0.196} & -0.221 \\
    & Annual Return \textuparrow & \textbf{0.914} & 0.122 & -0.213 & \underline{0.386} & -0.366 \\
     & Max Drawdown \textdownarrow & \underline{0.118} & \textbf{0.086} & 0.158 & 0.138 & 0.221 \\
     & Calmar Ratio \textuparrow & \textbf{7.778} & 1.429 & -1.344 & \underline{2.804} & -1.657 \\
    \midrule
    
    \multirow{4}{*}{NFLX} & Cumulative Returns \textuparrow & \textbf{0.323} & 0.129 & 0.142 & 0.027 & \underline{0.213} \\
    & Annual Return \textuparrow & \textbf{0.378} & 0.150 & 0.164 & 0.031 & \underline{0.248} \\
    & Max Drawdown \textdownarrow & 0.073 & 0.119 & \textbf{0.042} & 0.108 & \underline{0.051} \\
     & Calmar Ratio \textuparrow & \textbf{5.169} & 1.261 & 3.874 & 0.289 & \underline{4.811} \\
    \midrule
    
    \multirow{4}{*}{BABA} & Cumulative Returns \textuparrow & \textbf{0.119} & \underline{-0.167} & -0.204 & -0.231 & -0.297 \\
    & Annual Return \textuparrow & \textbf{0.126} & \underline{-0.175} & -0.214 & -0.242 & -0.311 \\
    & Max Drawdown \textdownarrow & \textbf{0.137} & 0.340 & 0.371 & 0.415 & \underline{0.301} \\
    & Calmar Ratio \textuparrow & \textbf{0.915} & \underline{-0.514} & -0.577 & -0.584 & -1.031 \\
    \midrule
    
    \multirow{4}{*}{PFE} & Cumulative Returns \textuparrow & \textbf{0.457} & -0.015 & 0.039 & -0.152 & \underline{0.068} \\
    & Annual Return \textuparrow & \textbf{0.502} & -0.016 & 0.042 & -0.163 & \underline{0.073} \\
    & Max Drawdown \textdownarrow & \underline{0.095} & 0.115 & \textbf{0.081} & 0.173 & 0.130 \\
    & Calmar Ratio \textuparrow & \textbf{5.273} & -0.137 & 0.515 & -0.942 & \underline{0.564} \\
    \midrule
    
    \multirow{4}{*}{DE} & Cumulative Returns \textuparrow & \textbf{0.513} & 0.129 & 0.040 & 0.211 & \underline{0.314} \\
    & Annual Return \textuparrow & \textbf{0.680} & 0.165 & 0.051 & 0.272 & \underline{0.411} \\
    & Max Drawdown \textdownarrow & \textbf{0.122} & 0.167 & 0.201 & 0.162 & \underline{0.146} \\
    & Calmar Ratio \textuparrow & \textbf{5.573} & 0.987 & 0.254 & 1.672 & \underline{2.823} \\
    \midrule
    
    \multirow{4}{*}{BHP} & Cumulative Returns \textuparrow & \textbf{0.256} & 0.140 & \underline{0.198} & -0.018 & 0.004 \\
    & Annual Return \textuparrow & \textbf{0.342} & 0.186 & \underline{0.264} & -0.024 & 0.005 \\
    & Max Drawdown \textdownarrow & \textbf{0.057} & \underline{0.077} & 0.223 & 0.147 & 0.223 \\
    & Calmar Ratio \textuparrow & \textbf{6.017} & \underline{2.422} & 1.184 & -0.161 & 0.024 \\
    \midrule
    
     \multirow{4}{*}{DUK} & Cumulative Returns \textuparrow & \textbf{0.370} & 0.048 & \underline{0.150} & 0.137 & 0.098 \\
    & Annual Return \textuparrow & \textbf{0.388} & 0.050 & \underline{0.157} & 0.143 & 0.103 \\
    & Max Drawdown \textdownarrow & \underline{0.023} & 0.055 & 0.043 & \textbf{0.019} & 0.054 \\
    & Calmar Ratio \textuparrow & \textbf{16.932} & 0.917 & 3.617 & \underline{7.514} & 1.916 \\
    \midrule
    \bottomrule
    \end{tabular}
}
\end{table}

%% file: pic_to_tex/bubble_pink_blue_plot.tex

\begin{figure}[!t]
    \centering
    \begin{subfigure}[t]{0.9\linewidth}
        \includegraphics[width=\textwidth]{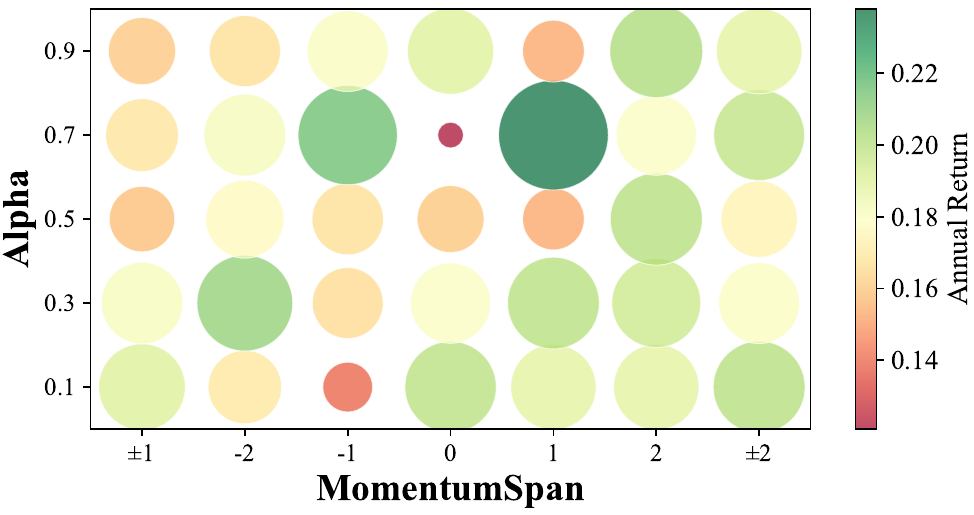}
        \label{fig:part_type_light}
    \end{subfigure}
    \caption{
       \textcolor{black}{\ours~under different combinations of hyperparameters. Larger bubbles mean higher performances.}
        \label{figs:bubble}
    }
\end{figure}

%% file: pic_to_tex/parameter_each.tex

\begin{figure}[!t]
	\centering
	\begin{subfigure}[t]{0.48\linewidth}
		\includegraphics[width=\textwidth]{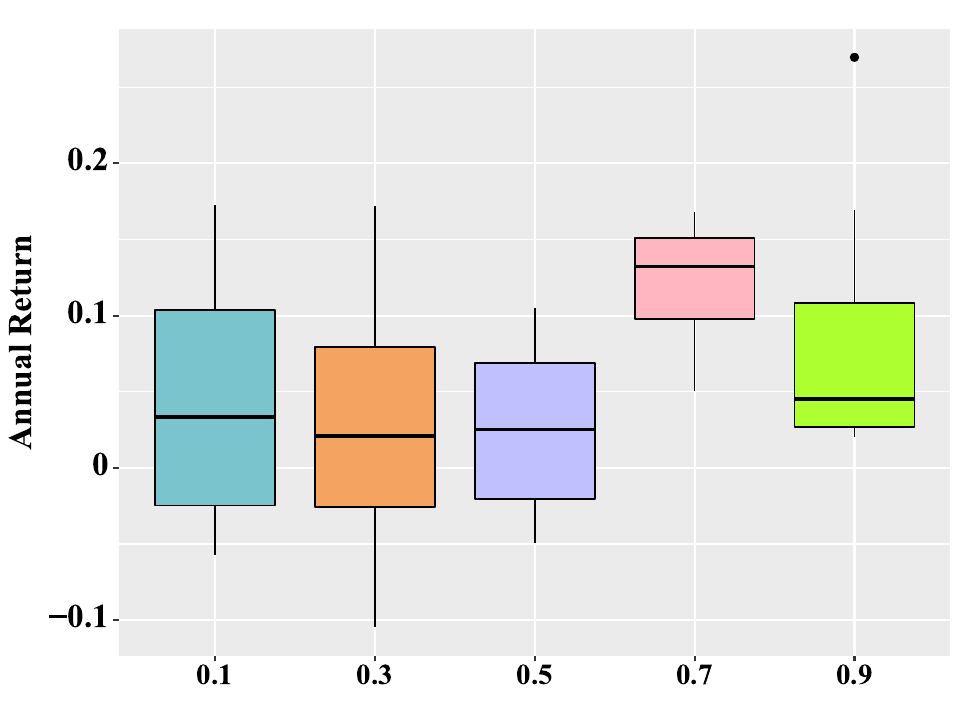}
		\label{parameter_boxplot_alpha_AAPL}
            \caption{$\alpha$ for AAPL}
	\end{subfigure}
	\begin{subfigure}[t]{0.48\linewidth}
		\includegraphics[width=\textwidth]{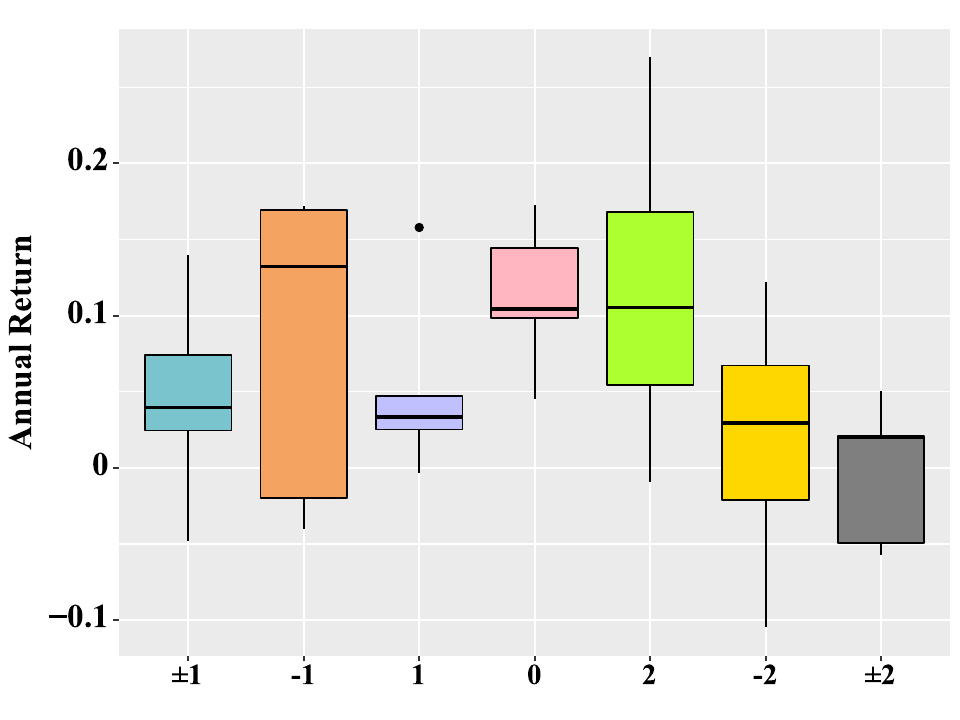}
	    \caption{$\Delta$ for AAPL}	
            \label{parameter_boxplot_Delta_AAPL}
	\end{subfigure}
    \begin{subfigure}[t]{0.48\linewidth}
		\includegraphics[width=\textwidth]{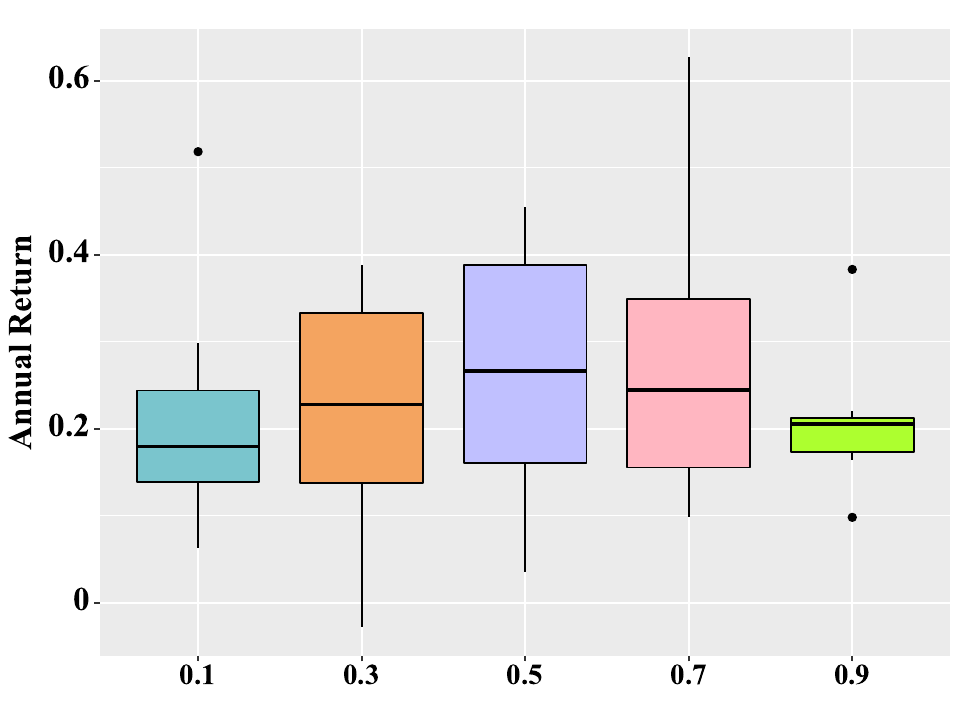}
            \caption{$\alpha$ for ECL}
		\label{parameter_boxplot_alpha_ECL}
	\end{subfigure}
    \begin{subfigure}[t]{0.48\linewidth}
		\includegraphics[width=\textwidth]{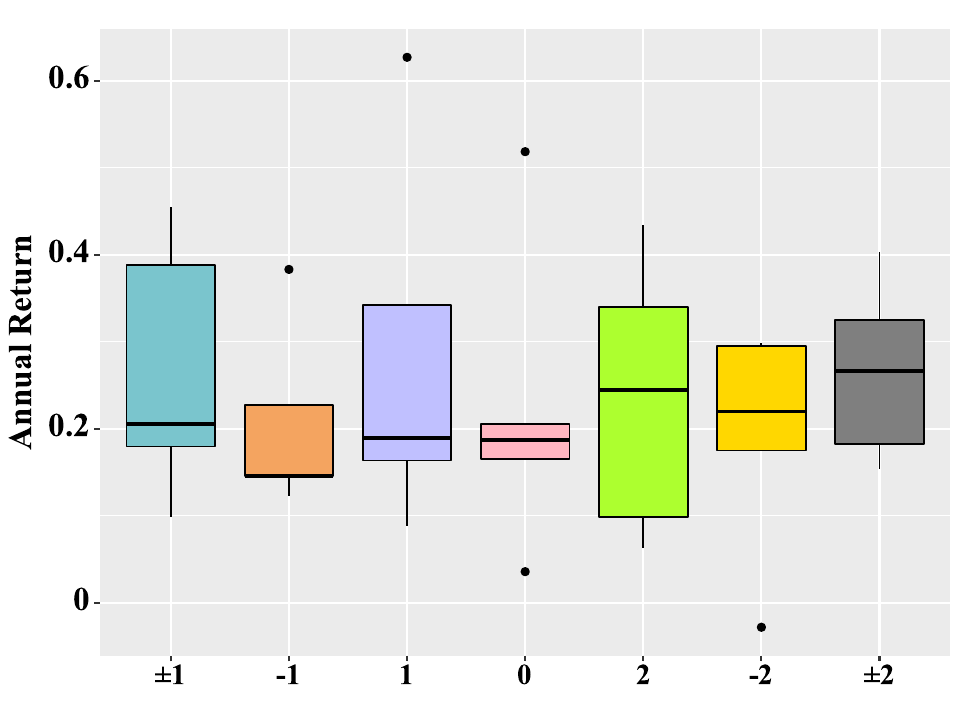}
            \caption{$\Delta$ for ECL}
		\label{parameter_boxplot_Delta_ECL}
	\end{subfigure}
	\caption{
       \textcolor{black}{The performance of the $\alpha$ and $\Delta$ on the Annual Return metric for AAPL and ECL stocks.}
        \label{figs:parameter_boxplot}
	}
\end{figure}

%% file: B-6-Discussion.tex
\section{Discussion - Case Study}
\label{sec:discussion}
Intuitively, initial biases, generated by bullish and bearish investors(bulls and bears), trigger price changes, and these biases gradually correct themselves over time, potentially leading to shifts in market behavior or trends. 
This reflects the irrational characteristics of the market and the significant role of investor sentiment in the formation of market prices. 
In the investor-driven market, individual companies may experience short-term biases due to fluctuations in investor sentiment or information asymmetry. 
On an industry level, the prices of entire sectors may exhibit similar irrational fluctuations, influenced by collective sentiment or macroeconomic factors.

\subsection{Cognitive Bias Initialization}
\input{pic_to_tex/radar_density}

This subsection explores biases initialization by formalizing the concepts of Attention Score (AS) and Bias, followed by an analysis of their distribution and industry-level effects, to quantify investors' attention to different topics and their divergence in interpretation within the stock market.
\begin{definition} 
    (Attention Score, AS) 
    For a given stock $s$, time window $\tau$, and topic $z$, we define the Attention Score of bulls and bears respectively as: $\mathit{AS}^{\mathit{s},\tau,\mathit{z}}_{\text{bull/bear}} = 1000 \cdot \left| \mathit{A}^{\mathit{s},\tau,\mathit{z}}_{\text{BU/BE}}\right|$, where $\mathit{A}_{\text{BU}}$ and $\mathit{A}_{\text{BE}}$ are the bullish and bearish attention weights derived from ~\ours~  model. The scaling factor of 1000 is applied for numerical readability. A higher $AS$ indicates stronger market focus on topic $z$ regarding stock $s$ at time $\tau$. 

    (Industry-level Attention Score, IAS)
    Correspondingly,  The aggregation of $\mathit{AS}$ for all stocks within a certain industry: $\mathit{IAS}^{\mathcal{I},\tau,z} = \sum_{\mathit{s} \in \mathcal{I}} (\mathit{AS}^{\mathit{s},\tau,\mathit{z}})$.
\end{definition}

The \textit{Attention Score} aims to measure the ``degree of attention'' or ``relevance''. Essentially it quantifies the market's focus on a particular stock at a specific time in relation to a specific topic.

\begin{definition}
     (Bias) 
     The absolute difference in Attention Scores between bulls and bears on a particular topic, to reflect their divergence in interpreting that topic: 
     $\mathit{Bias}^{\mathit{s},\tau,\mathit{z}} = \mathit{AS}^{\mathit{s},\tau,\mathit{z}}_{\text{bull}} - \mathit{AS}^{\mathit{s},\tau,\mathit{z}}_{\mathit{bear}}$, where $\mathit{AS}^{\mathit{s},\tau,\mathit{z}}_{\text{bull}}, \mathit{AS}^{\mathit{s},\tau,\mathit{z}}_{\text{bear}} \geq 0$.

    (Industry-level Bias, IBias)
    Correspondingly, $\mathit{IBias}^{\mathit{I},\tau,\mathit{z}} =\sum_{s\in \mathcal{I}} $ $\mathit{Bias}^{\mathit{s},\tau, \mathit{z}} $ $=\sum_{\mathit{s} \in \mathcal{I}} ( \mathit{AS}^{\mathit{s},\tau,\mathit{z}}_{\text{bull}} - \mathit{AS}^{\mathit{s},\tau,\mathit{z}}_{\text{bear}})$ $=\sum_{\mathit{s} \in \mathcal{I}} ( \mathit{AS}^{\mathit{s},\tau,\mathit{z}}_{\text{bull}}) - \sum_{\mathit{s} \in \mathcal{I}}(\mathit{AS}^{\mathit{s},\tau,\mathit{z}}_{\text{bear}})$$=\mathit{IAS}^{\mathit{I},\mathit{k}}_{\text{bull}}-\mathit{IAS}^{\mathit{I},\mathit{k}}_{\text{bear}}$ , where $\mathcal{I}$, $\tau$ and $\mathit{s}$ denotes a given industry, a time period and a set of stock in the industry $\mathcal{I}$.
\end{definition}

The bias metric shows significant market controversy. A large discrepancy in attention between bulls and bears indicates a substantial divergence in market understanding or predictions regarding that topic.

To analyze the performance of bull and bear from both stock and industry perspectives, we visualize the distribution of a specific stock, ABT, and a specific industry, technology, as shown in Fig. \ref{figs:radar_density}. 
This comparison provides valuable insights into the distinct yet complementary perspectives of market participants at the individual stock level (Micro) and broader industry level (Macro) behave, revealing unique patterns in attention and allocation strategies.

\par\noindent
\textbf{Micro Perspective}
\begin{itemize}[leftmargin=*]
    \item \emph{Attention Distribution Pattern}: Bulls and bears exhibit similar attention distribution patterns across different topics. For ABT stock, Topic1, Topic3, and Topic4 display long-tail distribution, while Topic2 approximates a Gaussian distribution. This indicates that certain topics (\textit{e.g.}, Topic2) receive consistent attention, whereas others (\textit{e.g.}, Topic3) generate significant divergence in market interpretation.  
    \item  \emph{Attention Allocation Differences}: Within the same topic, there are notable differences in how bulls and bears allocate attention. For instance, in Topic2, the distribution of $AS$ of bears exhibits kurtosis and displays a left skewness compared to that of bulls.
\end{itemize}

\par\noindent
\textbf{Macro Perspective}
\begin{itemize}[leftmargin=*]
    \item  \emph{Topic-Specific Attention}: Certain topics attract more attention from bulls or bears. For instance, shown as \ref{figs:radar_density}(right), $\mathit{IAS}_{\text{bull}}^{\text{Tech},\tau,\text{Z7}} > \mathit{IAS}_{\text{bull}}^{\text{Tech},\tau,\text{Z6}}$ indicates that bulls focus more on Z7 (\emph{Corporate News}) than on Z6 (Economic Growth). Bulls also exhibit heightened attention to Z10 (\emph{Consumer Demand}) and Z19 (\emph{Economic Policy Trends}), whereas bears show more interest on Z14 (\emph{Market Fluctuations}) and Z16 (\emph{Technology Stock Performance}).
    \item \emph{Topic Preference Differences}: Bulls and bears demonstrate significant differences in topic preferences. For example, $IBias^{Tech,t,Z6} > IBias^{Tech,t,Z7}$ suggests that the divergence in biases between bulls and bears is greater for Z6 (Economic Growth) than for Z7 (Corporate News). This reflects differences in how market signals are interpreted by various participants, highlighting the role of cognitive biases in shaping industry sentiment.  
\end{itemize}

\subsection{Bias Evolution}

Bias is not static - it evolves as investor attention shifts across time and topics. To characterize this process, we introduce the notion of \textit{Attention Migration} ($AM$). 

Let $(\tau,\mathit{z})$ and $(\tau',\mathit{z'})$ denote source and destination time-topic pairs. 
We define:
\begin{equation}
    \mathit{AM}^{\mathit{s},\tau\to \tau',\mathit{z}\to \mathit{z'}} = \mathit{AS}^{\mathit{s},\tau,\mathit{z}} \times \frac{\mathit{AS}^{\mathit{s},\tau',\mathit{z'}}}{\sum_{\mathit{z''}\in \mathcal{Z_{\tau'}}}{\mathit{AS}^{\mathit{s},\tau',\mathit{z''}}}}
\end{equation}

This measures the flow of attention from a topic $\mathit{z}$ at time $\tau$ to a future topic $\mathit{z'}$ at time $\tau'$, normalized by the relative salience of $\mathit{z'}$ among all topics at $\tau'$.
This formulation assumes that attention flows are both source-weighted and target-selective, i.e., higher migration occurs when both the origin is important and the destination is relatively more salient.

To quantify the divergence in attention shifts between opposing perspectives, we define the \textit{Bias Migration}: 
\begin{equation}
    \mathit{BM}^{\mathit{s},\tau\to \tau',\mathit{z}\to \mathit{z'}} = \mathit{AM}^{\mathit{s},\tau\to \tau',\mathit{z}\to \mathit{z'}}_{\text{bull}} -\mathit{AM}^{\mathit{s},\tau\to \tau',\mathit{z}\to \mathit{z'}}_{\text{bear}}
\end{equation}

It is important to note that the topic set may vary over time—new topics can emerge while others may disappear. This metric track the overall attention shifts and the evolving emotional divergence  between bulls and bears across time.

\par\noindent
\textbf{Micro Perspective: Stock-Level}

To illustrate the pattern of attention migration form a micro perspective, we analyze the attention data of ASML stock from September 2020 to December 2021, which is evenly divided into four time periods as an example. 

Fig. \ref{figs:migration_flow}(\textit{Left}) depicts the landscape of attention flow and biases shift over time. Each flow line represents the magnitude of $\mathit{AM}$, uncovering several important patterns:
(1) Attention flows as hot topics emerge or evolve.
(2) The attention to a topic at time $\tau$ is influenced by attention from the previous period. 
(3) Significant biases divergences between bull and bear at certain times intensifies market sentiment.

For example, Topic3 is dominant in the first periods, while Topic4 become more prominent later. Between Time1 to Time2, the increase in attention to Topic3 (bulls) and Topic2 (bears) is represented by the flow lines.

Fig. \ref{figs:migration_flow}(\textit{Right}) shows the volume of attention shifts, with red or green representing bullish and bearish biases, respectively. The former indicates that bulls pay more attention to a particular topic than bears, leading to an overall optimistic market sentiment towards that topic. Conversely, the latter signifies that bears focus more on a topic, resulting in a generally pessimistic market sentiment. 

For instance, regarding Topic2 of ASML, the market rapidly shifted from an extremely bullish bias to an extremely bearish bias, transitioning from a collective optimism to a collective pessimism. In contrast, for Topic3, the market maintained a bullish bias from Time1 to Time4, reflecting a sustained optimistic outlook.

\input{pic_to_tex/flow_hotmap}

\par\noindent
\textbf{Macro Perspective: Industrial-Level}

While $AM$ captures local transitions,  industry-level migration patterns reveal structural regularities. To capture this “magnetism” of topics within industries, we define the \textit{Industry Attention Migration} ($\mathit{IAM}$) toward a topic $z'$: 

\begin{equation}
    \mathit{IAM}^{\mathit{I},\mathit{z'}} =\sum_{\mathit{s} \in \mathcal{I}} \sum_{\mathit{z} \in \mathcal{Z}}\mathit{AM}^{\mathit{s},\tau \to \tau', \mathit{z} \to \mathit{z'}}  
\end{equation}

This measures how much total attention across time flows toward $z'$ within industry $\mathcal{I}$.

To normalize and compare across topics, we define the Attention Migration Propensity (AMP):
 \begin{equation}
      \mathit{AMP}^{\mathit{I},\mathit{z'}} = \frac{\mathit{IAM}^{\mathit{I},z'}}{\sum_{z'' \in \mathcal{Z}}{IAM^{\mathit{I},\mathit{z''}}}}
 \end{equation}

A high $\mathit{AMP}^{\mathit{I},\mathit{z'}}$ indicates that topic $\mathit{z'}$ frequently becomes the ``destination'' of attention reallocation within industry $\mathit{I}$, suggesting a directional tendency in how biases evolve.

As shown in Fig. \ref{figs:attention_flow} (\textit{Left}), each arc represents an industry (\emph{e.g.} I2, I6, I7, I8, I9, I11) or topic (Z1–Z10), while the chords represent the strength of $IAM$. Thicker chords signify stronger migration. 
For example, topics like Z1 (Energy Sector Dynamics) and Z3 (Vaccine) attract significant attention from industries I6 (Industrials) and I7 (Materials), suggesting these topics act as consistent focus centers. 
The proportion of flow at each industry node also indicates its topic selectivity ($AMP$). 
Industries like I6 show strong directional focus, channeling a large share of attention toward specific topics.

To examine the dynamics of bias, we define the Industry-level Bias Migration (IBM) toward topic $z'$ as the total difference in attention migration from bulls and bears across all topics and times, aggregated over an industry $\mathcal{I}$:
\begin{equation}
    IBM^{\mathit{I}, \mathit{z'}} = \sum_{\mathit{s} \in \mathcal{I}} \sum_{\tau \to \tau'} \sum_{\mathit{z} \in \mathcal{Z}_{\tau}} \left( AM^{\mathit{s}, \tau \to \tau', \mathit{z} \to \mathit{z'}}_{\text{bull}} - AM^{\mathit{s}, \tau \to \tau', \mathit{z} \to \mathit{z'}}_{\text{bear}} \right)
\end{equation}

This quantity captures how much more attention bullish investors are migrating toward topic $z'$, compared to bearish investors. A large positive value indicates directional convergence of bullish sentiment, whereas a large negative value implies increasing bearish focus.

To normalize these tendencies and enable comparison across topics, we define the Bias Migration Propensity (BMP):
\begin{equation}
    BMP^{\mathit{I}, \mathit{z'}} = \frac{\mathit{IBM}^{\mathit{I}, \mathit{z'}}}{\sum_{\mathit{z''} \in \mathcal{Z}} |IBM^{\mathit{I}, \mathit{z''}}|}
\end{equation}
Fig. \ref{figs:attention_flow} (\textit{Right}) displays a heatmap of the \textit{Bias Migration Propensity} ($\mathit{BMP}$), reflecting how bullish and bearish perspectives diverge across industries and topics. Green indicates topics where bulls dominate attention migration; red signals bear dominance.

\input{pic_to_tex/chord_hotmap}

Several patterns emerge: (1) Topic Z14 (Market Fluctuations) and Z16 (Tech Stocks Performance) show deep red rows across multiple industries (e.g., I4, I6, I9), indicating strong bearish focus, potentially linked to risk or negative sentiment.
(2) Z3 (IHS Markit score report) and Z6 (Schall Law Firm class action) are more green in industries like I3–I6, suggesting bullish narratives are spreading there.
(3) Some industries (e.g., I2, I7) remain neutral across most topics, showing balanced sentiment.
(4) These patterns reveal not only where attention is migrating but also the polarity of that migration. Strong industry-topic biases divergence can foreshadow directional moves in sector performance. 

The categorical encoding schemes utilized for the delineation of topics and industries are detailed in Appendix \ref{apx:Z_encoder} and \ref{apx:industry_encoder}.

\subsection{Market Behavior Adaptation} 
Through the above two steps, we have clearly observed that how attention forms and migrates. Then we will focus on how attention and bias drive trading decisions. We select data for AEP stock from September 2020 to July 2021 in Figure~\ref{fig:Behavioral}. The classification codes for topics are presented in Appendix \ref{apx:z_encoder}.

\begin{figure}[!h]
    \centering
    \includegraphics[width=1\linewidth]{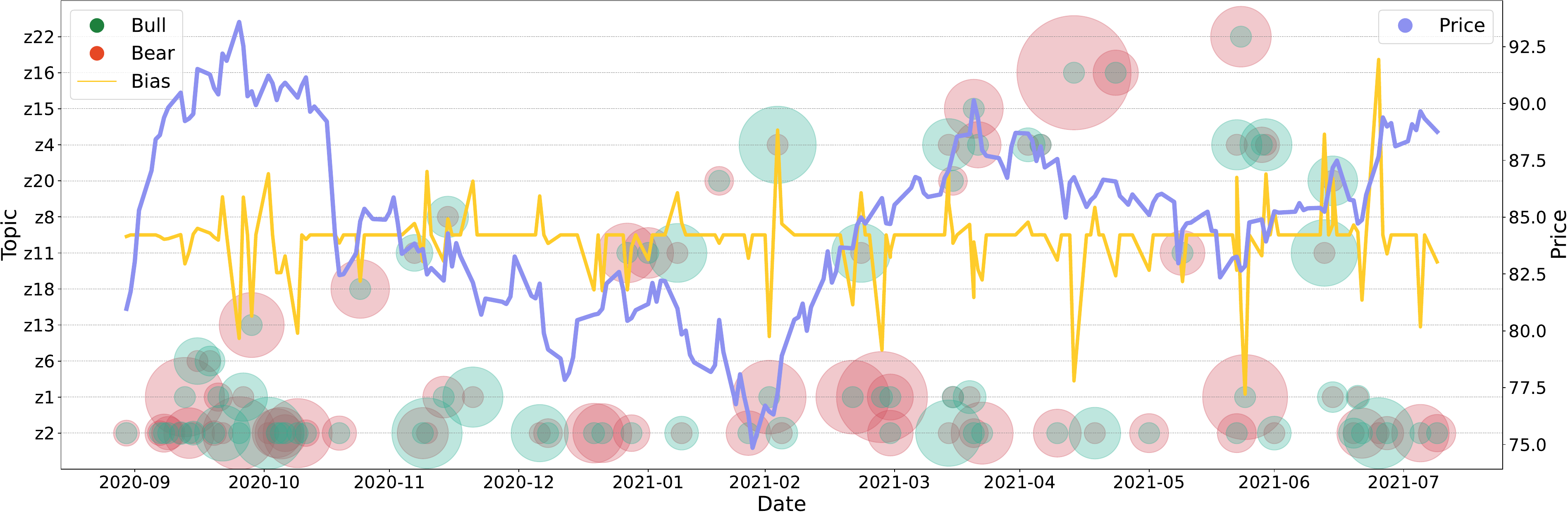}
    \caption{Bull-Bear Attention and Bias Migration over Time. Green bubbles indicate the intensity of bullish attention on each topic, while red bubbles represent bearish attention. The size of each bubble reflects attention magnitude, and their horizontal drift shows attention migration over time. The yellow line denotes the evolving bias, and the purple line shows the stock price.}
    \label{fig:Behavioral}
\end{figure}

\par\noindent
\textbf{Volatility under Symmetric but Intense Attention.}
In some cases, both bulls and bears focus intensely on the same high-stakes themes, such as lawsuits (z1, z2), resulting in a low net bias but a high total emotional intensity. This configuration leads to strong market participation on both sides, producing volatile price swings despite the absence of a dominant sentiment.

\begin{itemize}[leftmargin=*]
    \item \textbf{Case}: In October 2020, attention was highly concentrated on z1 (Investor lawsuits), z2 (Class action lawsuits), and z6 (Schall Law Firm class action). Bias oscillated sharply between positive and negative but centered near zero. Correspondingly, the price experienced a steep rally followed by a quick reversal.
\end{itemize}

\par\noindent
\textbf{Directional Trends Driven by Bias Polarization.}
When investor attention becomes structurally polarized—bulls focusing on constructive or fundamental themes (e.g., z4, z20), while bears remain anchored to pessimistic ones (e.g., z1, z2)—the bias increases in magnitude, providing a clear directional signal. These phases often align with trend continuation or reversals.

\begin{itemize}[leftmargin=*]
    \item \textbf{Case}: From February to April 2021, bulls shifted attention toward z4 (Energy Projects), z11 (Disability Inclusion), and z20  (Quarterly Earnings), while bears maintained focus on z1 (Investor lawsuits) and z2 (Class action lawsuits). The bias reached a local maximum, and the price steadily increased.
\end{itemize}

\par\noindent
\textbf{Sentiment Collapse Following Structural Realignment.}
Sudden shifts in topic focus—particularly when bearish attention expands to previously bullish domains—often signal emotional breakdowns and lead to price corrections. These structural realignments produce sharp transitions in the bias curve, often moving from peak optimism to extreme pessimism.

\begin{itemize}[leftmargin=*]
    \item \textbf{Case}: At the end of March 2021, bulls were optimistic about z2 (Class action lawsuits) and z4 (Energy projects), pushing prices to a local peak. However, bearish attention quickly spread to z4 (Energy projects) and z15 (Utility Portfolios), causing the bias to collapse and triggering a rapid price drop.
\end{itemize}

\par\noindent
\textbf{Market Recovery Triggered by Attention Rebalancing.}
Markets often recover when attention re-centers around forward-looking or solution-oriented themes, accompanied by renewed optimism from the bullish side. This rebalancing is typically marked by a reversal in the bias trend and a reallocation of attention to themes like z4 (Energy projects) or z20 (Quarterly Earnings Webcast).

\begin{itemize}[leftmargin=*]
    \item \textbf{Case}: In late May 2021, the price hit a low as bears focused on z1 (Investor lawsuits) and z22 (Customer Settlements), with bias reaching an extreme minimum. Soon after, bulls shifted attention toward z4 (Energy projects), and the price initiated a sustained upward rebound.
\end{itemize}

These patterns demonstrate that both the magnitude and structure of attention and bias—not just their directional values—are critical in explaining market behavior. Structural transitions in theme engagement often preempt major price inflection points, while emotionally balanced but intense phases produce unstable volatility. Recognizing and classifying these market states enables a deeper understanding of how information interpretation shapes price dynamics.

%% file: pic_to_tex/radar_density.tex

\begin{figure}[!h]
	\centering
	\begin{subfigure}[t]{0.48\linewidth}
		\includegraphics[width=\textwidth]{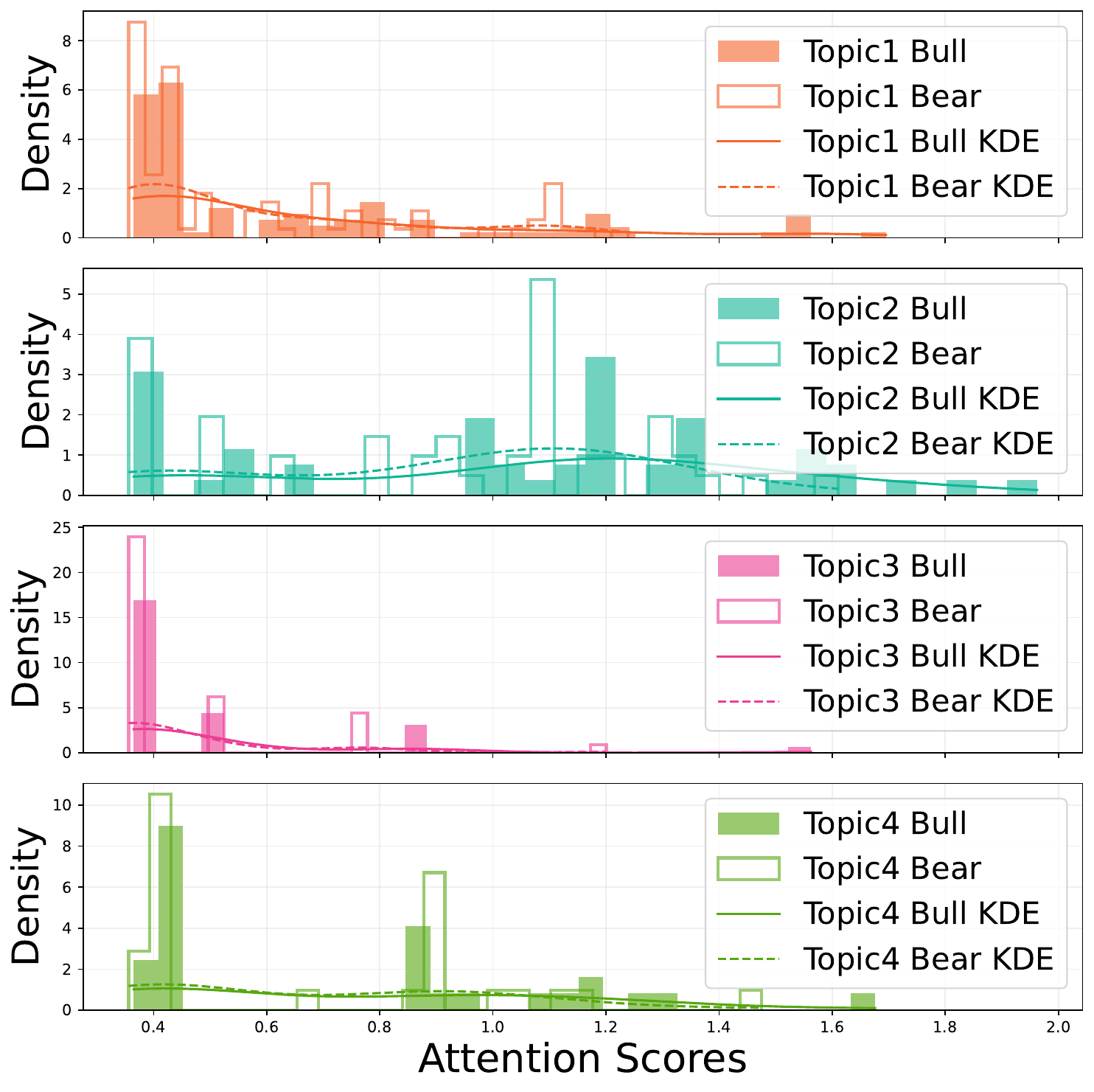}
		\label{parameter_boxplot_alpha_AAPL}
	\end{subfigure}
	\begin{subfigure}[t]{0.48\linewidth}
		\includegraphics[width=\textwidth]{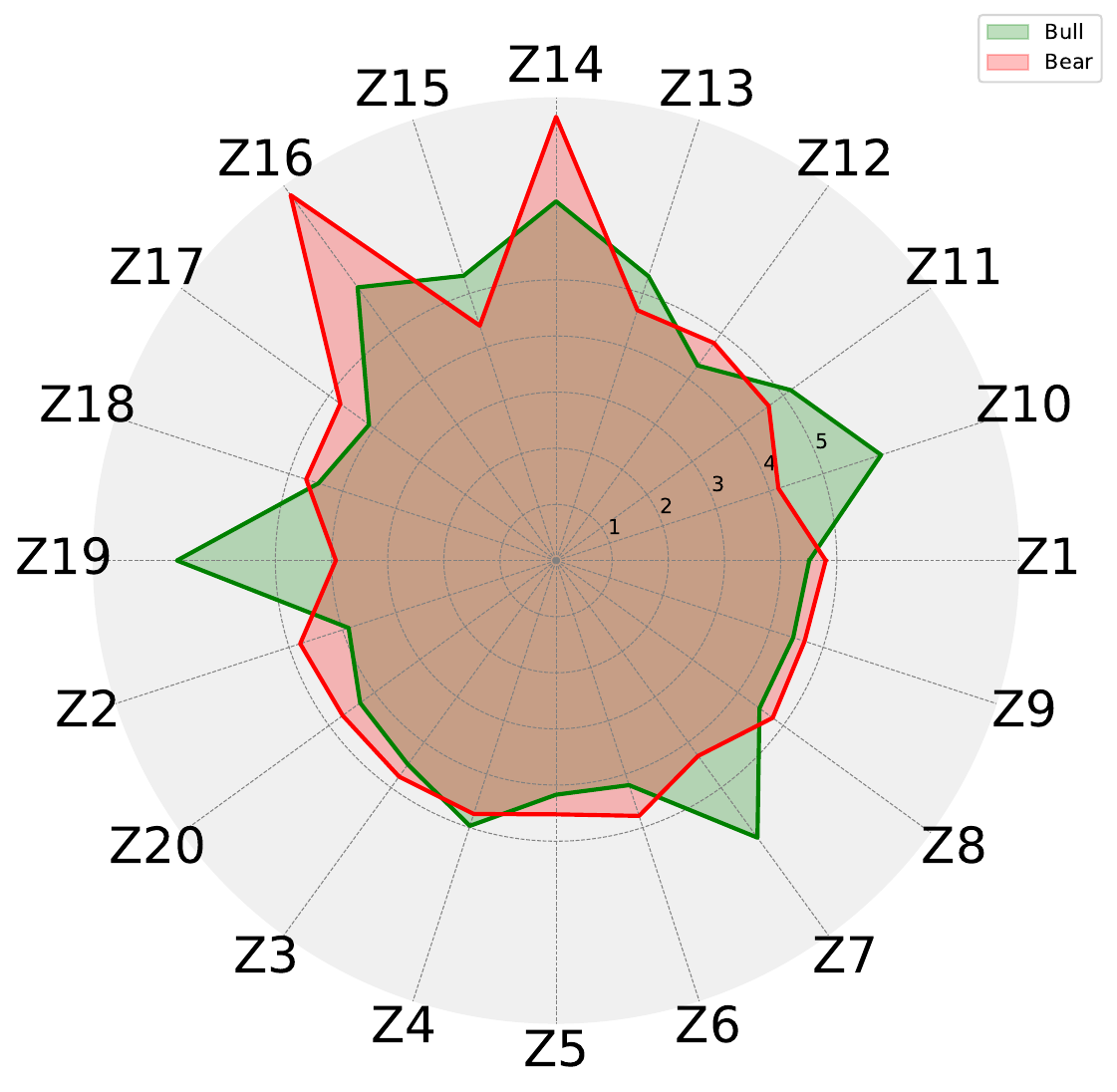}
            \label{parameter_boxplot_Delta_AAPL}
	\end{subfigure}
	\caption{
       \textcolor{black}{The bias of stock and Industry. \emph{Left}: Variations in attention patterns by $AS$ distribution between bullish and bearish investors.
       \emph{Right}: Differential attention, showed by $IAS^{Tech}$, to various topics within a specific industry (\textit{e.g.}, Technology Sector) across bullish and bearish investors.}
        \label{figs:radar_density}
	}
\end{figure}

%% file: pic_to_tex/flow_hotmap.tex

\begin{figure}[!h]
	\centering
	\begin{subfigure}[t]{0.6\linewidth}
		\includegraphics[width=\textwidth]{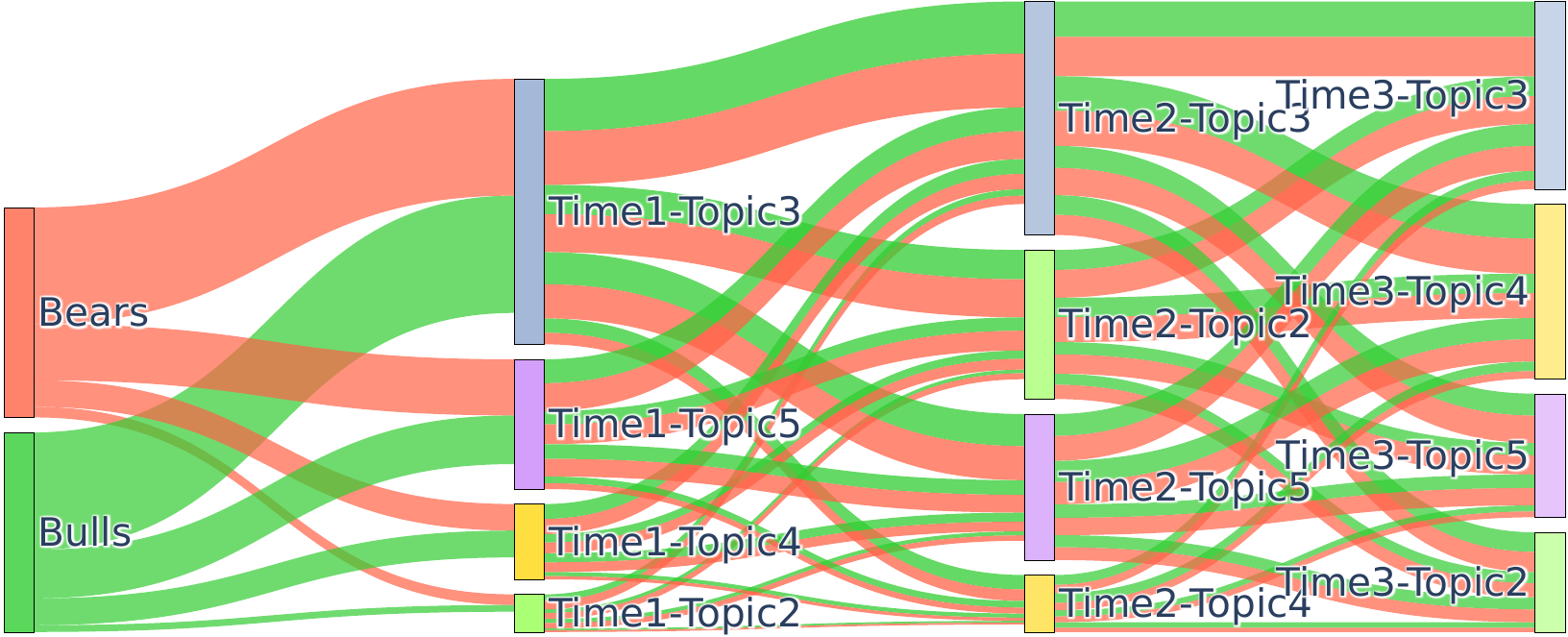}
            \label{fig:migration}
	\end{subfigure}
	\begin{subfigure}[t]{0.38\linewidth}
		\includegraphics[width=\textwidth]{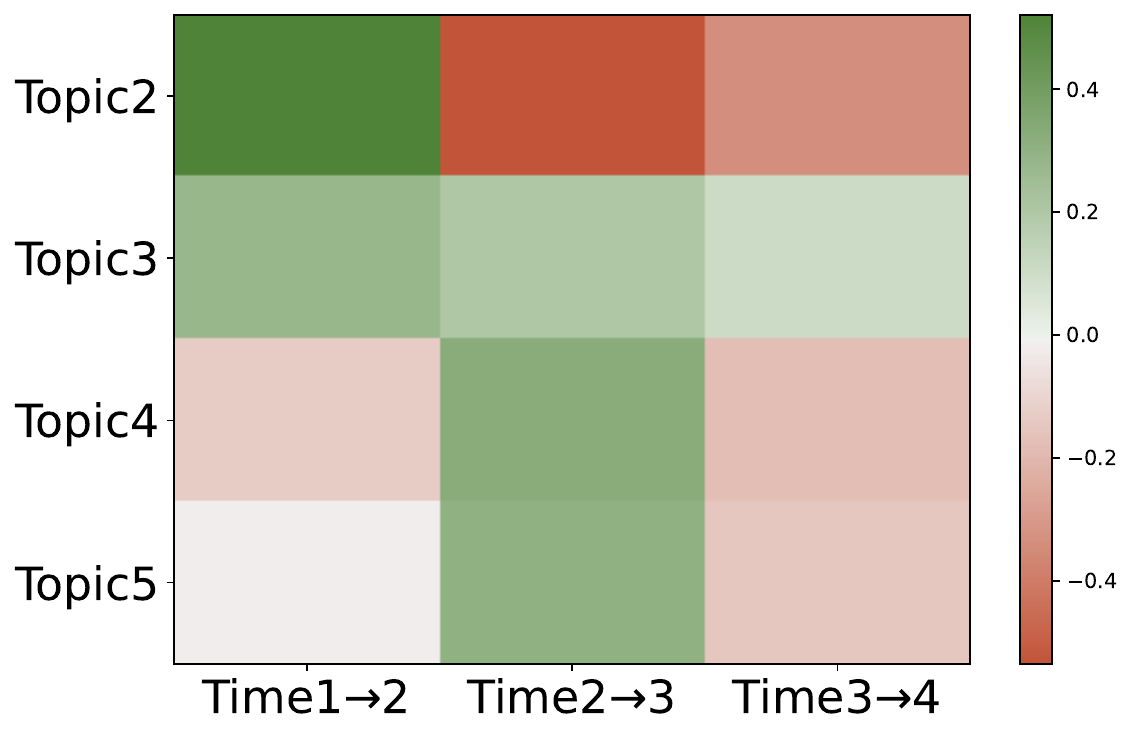}	
            \label{fig:hot_atten}
	\end{subfigure}
	\caption{
       \textcolor{black}{\emph{Left}: The attention migration across topics over time, highlighting attention flow. Flow thickness shows $AM$. \emph{Right}: The magnitude of bias shift across different time periods and various topics, showing $BM$.}
        \label{figs:migration_flow}
	}
\end{figure}

%% file: pic_to_tex/chord_hotmap.tex

\begin{figure}[!h]
	\centering
	\begin{subfigure}[t]{0.45\linewidth}
		\includegraphics[width=\textwidth]{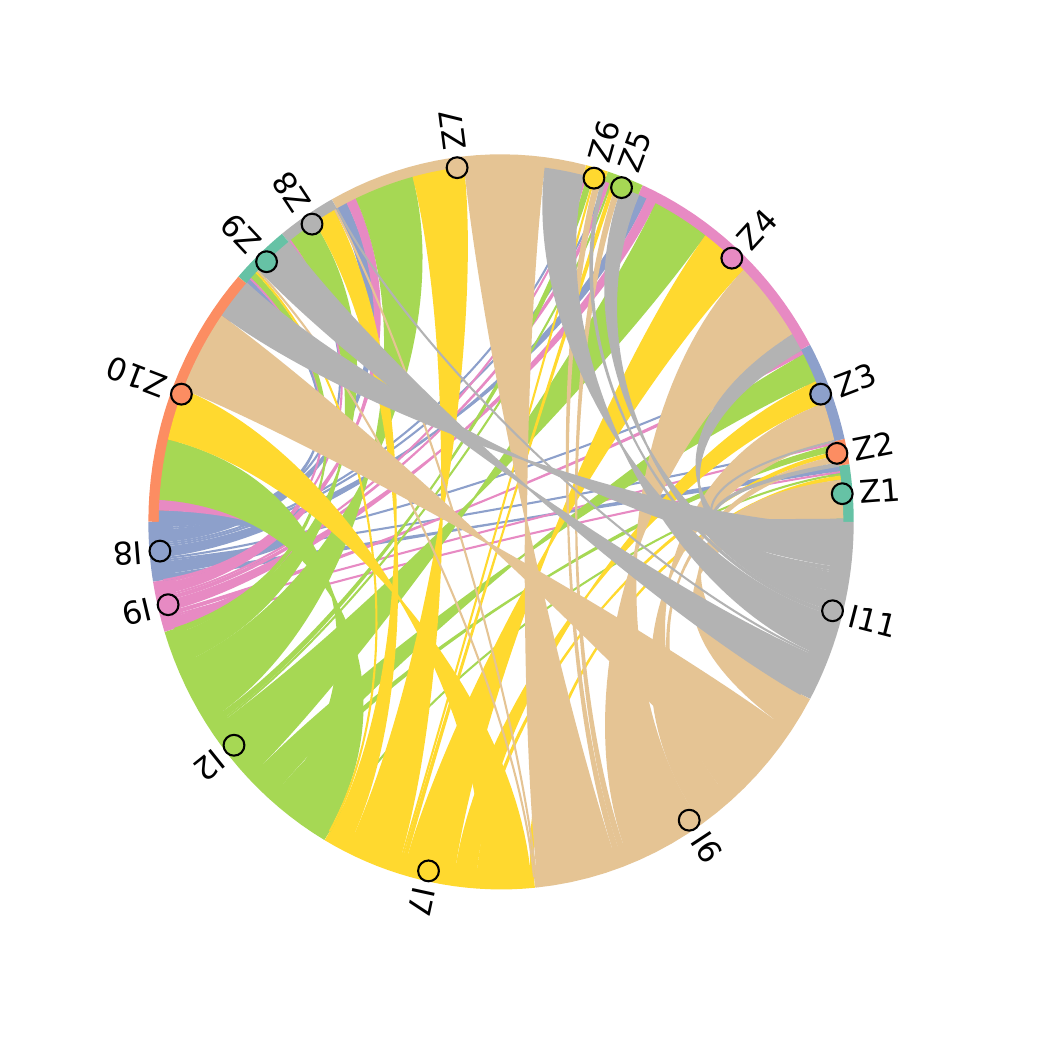}
            \label{fig:chord}
	\end{subfigure}
	\begin{subfigure}[t]{0.48\linewidth}
		\includegraphics[width=\textwidth]{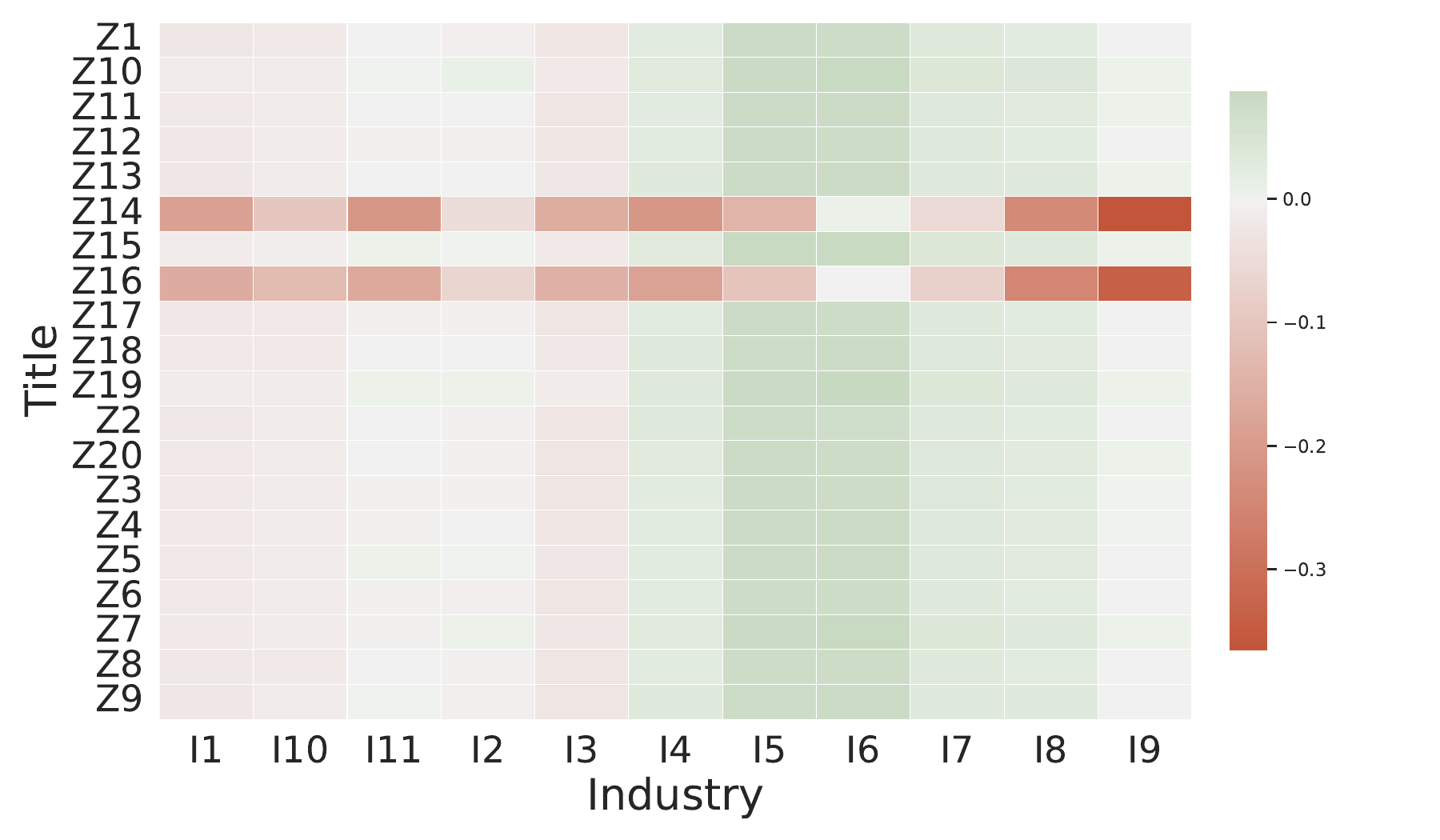}	
            \label{fig:hot_bias}
	\end{subfigure}
	\caption{
       \textcolor{black}{\emph{Left}: Flow thickness by $IAM$, the proportion of the flow at the industry node indicates $AMP$. 
    \emph{Right}: Bull-Bear Industry-Specific Perspective Difference. $BMP$ determines the intensity of colors in heatmap.}
        \label{figs:attention_flow}
	}
\end{figure}

%% file: B-7-Conclusion.tex
\section{Conclusion}
\label{sec:Conclusion}
This paper introduces a behavioral framework for modeling market regimes, grounded in the dynamics of bull and bear sentiment. By aligning investor-driven behaviors with evolving cognitive biases, we reveal how these biases emerge and manifest in observable market patterns. Distinct from prior approaches focused solely on pattern extraction, our model captures the competitive interplay between opposing investor stances, offering both interpretability and predictive power. Extensive experiments on 110 public datasets across 10 industries validate the framework’s robustness and broad applicability.

Looking forward, several promising directions remain. These include enhancing B4’s temporal granularity via real-time sentiment from high-frequency news and social media, extending the dual-bias modeling to multi-asset portfolios to capture cross-asset sentiment interactions, and applying the framework to derivative pricing or systemic risk scenarios. These extensions aim to further bridge behavioral finance theory with actionable machine learning systems in real-world financial environments.

%% file: D-Appendix/D-8-Appendix-Variables.tex
\section{Summary of Variables} 
\label{variables}

\textbf{Table~\ref{tab:variable_summary}} summarizes all notations used in our bias-attention modeling, covering both stock-level and industry-level constructs. These variables underpin the quantitative formulation of attention scores, bias strength, and their temporal migration dynamics across topics and market entities.

\begin{table*}[htbp]
\centering
\caption{Summary of Variables in Attention and Bias Modeling}
\label{tab:variable_summary}
\renewcommand{\arraystretch}{0.9}  
\resizebox{\textwidth}{!}{
    \begin{tabular}{c|l|l}
    \toprule
    \midrule
    \textbf{Symbol} & \textbf{Definition} & \textbf{Description} \\
    \midrule
    $\mathbf{X}_\mathit{t}^\mathit{P}$ & $[\mathit{x_t}^{\text{Open}}, \mathit{x_t}^{\text{Close}}, \mathit{x_t}^{\text{Low}}, \mathit{x_t}^{\text{High}}]$ & Multivariate price input at time $t$ \\
    \midrule
    $\delta$ & Sliding window length & Historical window length for time-series input \\
    \midrule
    $\mathit{z}$ & Topic index & Index of a topic (e.g., Corporate News, Tech Fluctuations) \\
    \midrule
    $\tau$ & Time window & Discrete time period or interval \\
    \midrule
    $s$ & Stock symbol & A specific stock in the market \\
    \midrule
    $\mathit{I}$ & Industry type & A specific industry, e.g., finance, energy, or healthcare \\
    \midrule
    $\mathcal{I}$ & Industry set & A collection of stock in industry $\mathit{I}$, i.e., $\mathcal{I} = \{ \mathit{s}_1, \mathit{s}_2, \dots \}$ \\
    \midrule
    $\mathcal{Z}_\tau$ & Topic set at $\tau$ & Topics available at time $\tau$ \\
    \midrule
    $\textit{A}_{\text{BU}}^{s,\tau,\mathit{z}}, \mathit{A}_{\text{BE}}^{\mathit{s},\tau,\mathit{z}}$ & -- & Bearish attention weight (output of our model) \\
    \midrule
    $AS_{\text{bull/bear}}^{\mathit{s},\tau,\mathit{z}}$ & $1000 \cdot |\mathit{A}_{\text{BU/BE}}^{\mathit{s},\tau,\mathit{z}}|$ & Scaled attention score for bulls/bears \\
    \midrule
    $\mathit{IAS}^{\mathit{I},\tau,\mathit{z}}$ & $\sum_{\mathit{s} \in \mathcal{I}} \mathit{AS}^{\mathit{s},\tau,\mathit{z}}$ & Industry-level attention score \\
    \midrule
    $\mathit{Bias}^{\mathit{s},\tau,\mathit{z}}$ & $\mathit{AS}_{\text{bull}}^{\mathit{s},\tau,\mathit{z}} - \mathit{AS}_{\text{bear}}^{\mathit{s},\tau,\mathit{z}}$ & Bias score at stock level \\
    \midrule
    $\mathit{IBias}^{\mathit{I},\tau,\mathit{z}}$ & $\mathit{IAS}_{\text{bull}} - \mathit{IAS}_{\text{bear}}$ & Industry-level bias score \\
    \midrule
    $\mathit{AM}^{\mathit{s},\tau \to \tau',\mathit{z} \to \mathit{z'}}$ & $\mathit{AS}^{\mathit{s},\tau,\mathit{z}} \cdot \frac{\mathit{AS}^{\mathit{s},\tau',\mathit{z'}}}{\sum_{\mathit{z''}} \mathit{AS}^{\mathit{s},\tau',\mathit{z''}}}$ & Attention migration score across time and topics \\
    \midrule
    $\mathit{BM}^{\mathit{s},\tau \to \tau',\mathit{z} \to \mathit{z'}}$ & $\mathit{AM}_{\text{bull}} - \mathit{AM}_{\text{bear}}$ & Bias migration at stock level \\
    \midrule
    $\mathit{IAM}^{\mathit{I}, \mathit{z'}}$ & $\mathit{IAM}^{\mathit{I},\mathit{z'}} =\sum_{\mathit{s} \in \mathcal{I}} \sum_{\mathit{z} \in \mathcal{Z}}\mathit{AM}^{\mathit{s},\tau \to \tau', \mathit{z} \to \mathit{z'}}  $ & Total industry attention inflow to topic $\mathit{z'}$ \\
    \midrule
    $\mathit{AMP}^{\mathit{I}, \mathit{z'}}$ & $\mathit{AMP}^{\mathit{I},\mathit{z'}} = \frac{\mathit{IAM}^{\mathit{I},z'}}{\sum_{z'' \in \mathcal{Z}}{IAM^{\mathit{I},\mathit{z''}}}}$ & Normalized attention migration propensity \\
    \midrule
    $\mathit{IBM}^{\mathit{I}, \mathit{z'}}$ & $IBM^{\mathit{I}, \mathit{z'}} = \sum_{\mathit{s} \in \mathcal{I}} \sum_{\tau \to \tau'} \sum_{\mathit{z} \in \mathcal{Z}_{\tau}} \left( AM^{\mathit{s}, \tau \to \tau', \mathit{z} \to \mathit{z'}}_{\text{bull}} - AM^{\mathit{s}, \tau \to \tau', \mathit{z} \to \mathit{z'}}_{\text{bear}} \right)$ & Net industry-level bias migration \\
    \midrule
    $BMP^{\mathit{I}, \mathit{z'}} $ & $\frac{\mathit{IBM}^{\mathit{I}, \mathit{z'}}}{\sum_{\mathit{z''} \in \mathcal{Z}} |IBM^{\mathit{I}, \mathit{z''}}|}$ & Normalized bias migration propensity \\
    \midrule
    \bottomrule
    \end{tabular}
}
\end{table*}

%% file: D-Appendix/D-2-Appendix-Datasets.tex
\section{Dataset Details}\label{datasets}
The following tables provide a comprehensive overview of industry-related topics, their corresponding codes, and the stock classifications from the CMIN dataset. These tables are structured to facilitate the analysis of thematic and industry-specific factors that influence stock performance and market dynamics. To understand how these thematic developments correlate with stock behaviors, we use topic modeling techniques, specifically BERTopic~\cite{grootendorst2022bertopicneuraltopicmodeling} and LDA~\cite{lda2003}, to capture the underlying themes and sentiment shifts.

The choice of BERTopic and LDA as topic modeling tools reflects the specific challenges posed by the dataset, where BERTopic is used for capturing detailed, semantic themes from stock-specific topics, and LDA is employed for handling large-scale industry-level data where fast processing is critical.

\begin{itemize}[leftmargin=*]
    \item Table \ref{apx:Z_encoder} presents a different set of thematic codes (Z1–Z20), modeled using LDA, focusing on broader economic and sector-specific issues, including energy dynamics, mergers and acquisitions, tech industry trends, and market fluctuations.
    \item Table \ref{apx:z_encoder} introduces a set of industry topics (z1–z22), derived using BERTopic, covering key areas such as legal actions, investment advice, earnings reports, and market sentiment, which reflect various factors impacting industry sectors.
    \item Table \ref{apx:industry_encoder} maps industry codes (I1–I10) to their respective explanations and lists major companies within each sector, such as technology, healthcare, and financial services, providing a clearer understanding of industry composition and stock categorization.
\end{itemize}

These tables serve as a foundation for analyzing how thematic developments correlate with stock behaviors across different industries, aiding in investment decision-making and risk assessment strategies.

\begin{table}[htbp]
\centering
\caption{Industry topic codes and explanations (\emph{cf.} Fig. \ref{fig:radar}).}
\label{apx:Z_encoder}
\renewcommand{\arraystretch}{1.1}
\resizebox{0.8\columnwidth}{!}{%
    \begin{tabular}{c|p{6cm}}
    \toprule
    \midrule
      \textbf{Topic} &  \textbf{Explanation} \\

      \midrule
      Z1 & Energy Sector Dynamics \\
      \hline
      Z2 & Investment and Funds \\
      \hline
      Z3 & Vaccine	 \\
      \hline
      Z4 & Executive Changes \\
      \hline
      Z5 & Investment Strategies	 \\
      \hline
      Z6 & Economic Growth \\
      \hline
      Z7 & Corporate News \\
      \hline
      Z8 & Pharmaceutical and Regulatory	 \\
      \hline
      Z9 & Tech Industry \\
      \hline
      Z10 & Consumer Demand \\
      \hline
      Z11 & E-Commerce and Data \\
      \hline
      Z12 & Earnings Reports	 \\
      \hline
      Z13 & Mergers and Acquisitions		 \\
      \hline
      Z14 & Market Fluctuations		 \\
      \hline
      Z15 & Pharmaceutical Companies	 \\
      \hline
      Z16 & Tech Stocks Performance		 \\
      \hline
      Z17 & Industry Developments	 \\
      \hline
      Z18 & Company Dynamics		 \\
      \hline
      Z19 & Economic Policy Trends		 \\
      \hline
      Z20 & Economic Cycles		 \\
    \midrule
    \bottomrule
    \end{tabular}
}
\end{table}

\begin{table}[htbp]
\centering
\caption{AEP topic codes and explanations (\emph{cf.} Figs. \ref{fig:desity} and \ref{fig:Behavioral}.)}
\label{apx:z_encoder}
\renewcommand{\arraystretch}{1.1}
\scalebox{.90}{
    \begin{tabular}{c|p{6cm}}
    \toprule
    \midrule
    \textbf{Topic} &  \textbf{Explanation} \\ 
    \midrule
      z1 & Investor lawsuits \\
      \hline
      z2 & Class action lawsuits \\
      \hline
      z3 & IHS Markit score report \\
      \hline
      z4 & Energy projects \\
      \hline
      z5 & Utilities Sector Appraise \\
      \hline
      z6 & Schall Law Firm class action \\
      \hline
      z7 & Moody's ratings and outlook \\
      \hline
      z8 & Investment advice \\
      \hline
      z9 & Over-selling \\
      \hline
      z10 & Law lawsuits \\
      \hline
      z11 & Disability Inclusion \\
      \hline
      z12 & Earnings and Revenue Estimates \\
      \hline
      z13 & ETF flows \\
      \hline
      z14 & Analysts' Views \\
      \hline
      z15 & Durable Utility Stocks Portfolios \\
      \hline
      z16 & Bids for Coal Projects \\
      \hline
      z17 & Ratings and Outlook \\
      \hline
      z18 & Analyst blog highlights \\
      \hline
      z19 & Value comparison with DTE \\
      \hline
      z20 & Quarterly Earnings Webcast \\
      \hline
      z21 & Texas Power Demand Hits Record \\
      \hline
      z22 & Customers Receive Settlement \\
    \midrule
    \bottomrule
\end{tabular}
}
\end{table}
      

\begin{table}[!t]
\centering
\caption{Industry codes and explanations, and corresponding stocks from the CMIN dataset.}
\label{apx:industry_encoder}
\setlength{\tabcolsep}{2.5pt}
\renewcommand{\arraystretch}{1.1}
\scalebox{.82}{
    \begin{tabular}{c|>{\centering\arraybackslash}m{2.5cm}|p{6.5cm}}
    \toprule
    \midrule
      \textbf{Industry} &  \textbf{Explanation} & \textbf{Stocks} \\ 
      \midrule
      I1 & Communication & CHTR, CMCSA, NFLX, T, TMUS, VZ\\
      \hline
      I2 & Consumer& AMZN, BABA, DIS, JD, LOW, MCD, NFLX, NKE, SBUX, TGT, COST, DEO, KO, PG, PM, UL, WMT\\
      \hline
      I3 & Energy	& BP, COP, CVX, EQNR, RDS-B, XOM, TTE, PTR \\
      \hline
      I4 & Financial & ACN, ADP, AC, BRK-A, JPM, MA, SCHW, V \\
      \hline
      I5 & Healthcare & ABBV, ABT, JNJ, LLY, NVO, NVS, PFE, TMO, UNH \\
      \hline
      I6 & Industrials & BA, CAT, DE, GE, HON, MMM, RTX, UPS \\
      \hline
      I7 & Materials	& APD, BHP, ECL, FCX, RIO, SHW, VALE \\
      \hline
      I8 & Real Estate & AMT, CCI, DLR, EQIX, O, PLD, PSA, SBAC\\
      \hline
      I9 & Technology & AAPL, ADBE, ASML, AVGO, CSCO, FB, GOOG, MS, NVDA, ORCL, PYPL, TSM\\
      \hline
      I10 & Utilities & AEP, DUK, EXC, NEE, NGG, SO, SRE, XEL\\
    \midrule
    \bottomrule
    \end{tabular}
}
\end{table}

%% file: D-Appendix/D-7-Appendix-Algorithm.tex
\section{Algorithm design and analysis} 
\label{algo}

\begin{algorithm}[htbp]
    \caption{\ours}
    \label{algorithm:masp}
    \SetKwInOut{Input}{Input}\SetKwInOut{Output}{Output}
    \Input{Price sequence $X^{P}_{t} = \{x_t^{(Open)},x_t^{(Close)},x_t^{(Low)},x_t^{(High)}\}^{t}_{t-\delta }$, News corpus $X^N_{t} = \{x_t^N\}_{t-\delta}^t$, Learning rate $\eta$, Epochs $N$, Trade-off coefficient $\alpha$, MomentumSpan $\Delta$, Pretrained embeddings $H$, Price2Concept Layer $\text{P2C}$, Bias Simulation Layers $\text{BS}$}
    \Output{Prediction $Y_{t}'$, Parameters $\theta$} 
    \textbf{Initialize} $\theta\sim \mathcal{N}(0, 0.01)$;
    
    \For{$\forall n \in \{1,...,N\}$}{
        $\mathbf{H'} \gets \text{Linear}(\mathbf{H})$;
        
        $\mathbf{E}_{\text{price}} \gets \text{P2C}(\mathbf{X}_{\mathit{t}}^{\mathit{P}}; \textbf{H}')$ via Eq.\eqref{P2C};
        
        $\mathbf{X}^{\text{Aug}}_{t} \gets \text{Augment}(\mathbf{X}_{t}^{N})$ via  Eq.\eqref{aug};

        $\mathbf{E}_{\text{price}} \gets \text{Embed}(\mathbf{X}^{\text{Aug}}_{t})$;
        
        $\mathbf{E} \gets \text{Concat}(\mathbf{E}_{\text{text}}, \mathbf{E}_{\text{price}})$; 
        
        
        $\mathbf{h}_{\text{BU}}, \mathbf{h}_{\text{BE}} \gets \text{BS}(\mathbf{E})$ via {Eq.\eqref{bb_index}};
        
        
        $y_{t} \gets \text{Label}(\mathbf{X}_{t}^{P})$ via {Eq.\eqref{mtl}};
        
        $\mathcal{P}, \mathcal{N} \gets \emptyset$\;
        \For{$\forall \delta \in \Delta$}{
            $\mathcal{P} \gets \mathcal{P} \cup \{x_{j}| y_i = y_j, |i - j| \leq \delta\}$;
            
            $\mathcal{N} \gets \mathcal{N} \cup \{x_{j}| y_i \neq y_j, |i - j| \leq \delta\}$; 
        }
        
        $\text{Compute }  \mathcal{L}_{\text{Comp}},\mathcal{L}_{\text{Mar}}$ via {Eq.\eqref{dual}};
        
        $\text{Compute } \mathcal{L}_{\text{CE}}$ via {Eq.\eqref{ce}};
        
        $\mathcal{L}_{\text{Total}}=\alpha \cdot (\mathcal{L}_{\text{Comp}}+ \mathcal{L}_{\text{Mar}})+(1-\alpha)\cdot \mathcal{L}_{\text{CE}}$;
        
        $\theta \gets \text{Adam}(\mathcal{L}_{\text{Total}}, \eta)$;
    }
\end{algorithm}

Algorithm \ref{algorithm:masp} present the main steps of \ours\ approach.%
\par\noindent
\textbf{Space Complexity:} The algorithm's memory footprint is dominated by the storage of sequential embeddings and intermediate feature maps. The Price-to-Concept (P2C) layer requires $O(\delta \times d)$ space to store price embeddings, where $\delta$ is the temporal window size and $d$ denotes the embedding dimension. The concatenated feature matrix $\mathbf{E}$ consumes $O(\delta \times (d + k))$ space, with $k$ representing the augmented text feature dimension. Momentum-based contrastive learning introduces $O(|\Delta| \times \delta^2)$ space complexity for storing positive/negative pairs $\mathbf{P}$ and $\mathbf{N}$, where $|\Delta|$ is the cardinality of momentum spans. Parameter storage $\theta$ contributes $O(m)$, where $m$ is the total trainable parameters. The overall space complexity scales linearly with $\delta$ and quadratically with $|\Delta|$.

\par\noindent
\textbf{Time Complexity:} Dominated by iterative embedding transformations and pair-wise comparisons, the P2C projection and Bias Simulation (BS) layers incur $O(N \times \delta \times d^2)$ operations per epoch due to dense matrix multiplications, where $N$ is the number of epochs. Momentum pair construction requires $O(N \times |\Delta| \times \delta^2)$ time for sliding window comparisons across $\delta$-length sequences. The Adam optimizer contributes $O(N \times m)$ complexity for parameter updates. The total time complexity is $O(N \times (\delta d^2 + |\Delta|\delta^2 + m))$, demonstrating polynomial scaling with respect to sequence length and momentum span configurations.

%% file: D-Appendix/D-6-Appendix-Results.tex
\input{tables/appendix_industry_compare}

\begin{figure*}[!t]
	\centering
        \begin{subfigure}[t]{0.9\linewidth}
		\includegraphics[width=\textwidth]{picture/lineplot_new/legend.pdf}
		\label{fig:clr_ml_light}
	\end{subfigure}
	\begin{subfigure}[t]{0.24\linewidth}
		\includegraphics[width=\textwidth]{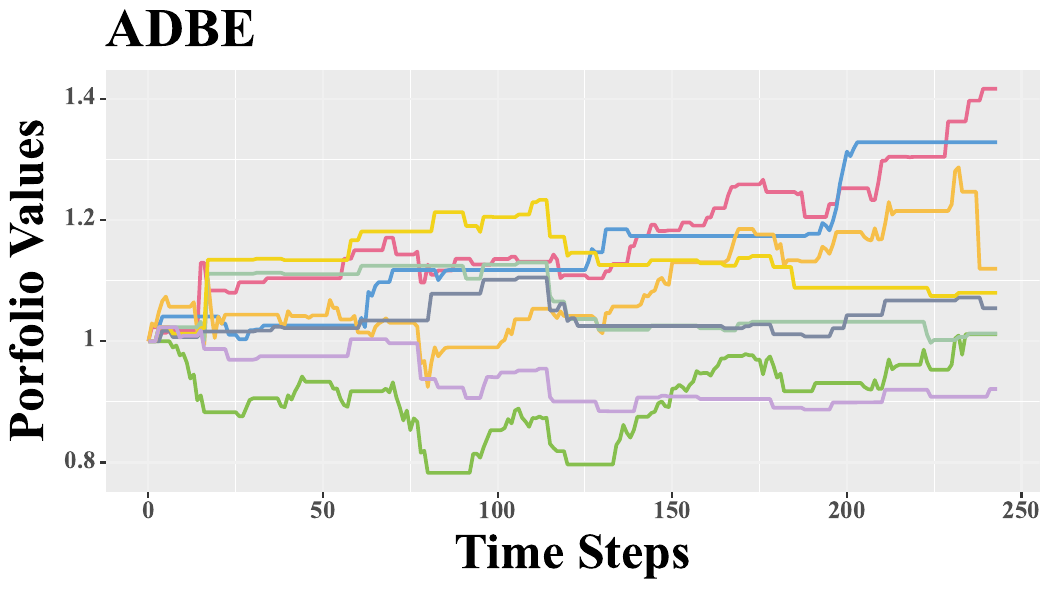}
		\label{fig:clr_ml_light}
	\end{subfigure}
	\begin{subfigure}[t]{0.24\linewidth}
		\includegraphics[width=\textwidth]{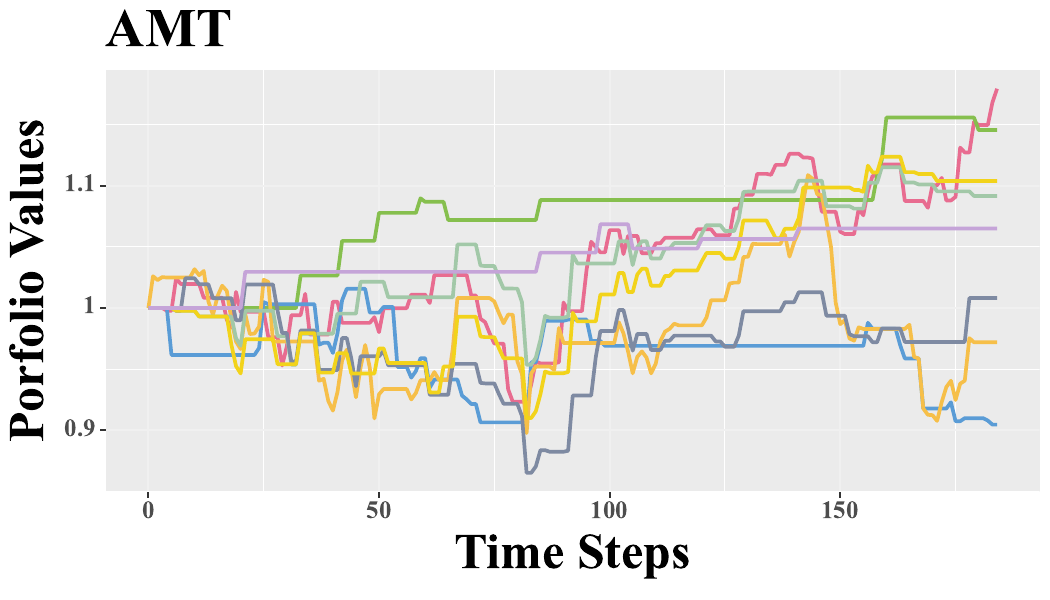}
		\label{fig:clr_ml_wmf}
	\end{subfigure}
	\begin{subfigure}[t]{0.24\linewidth}
		\includegraphics[width=\textwidth]{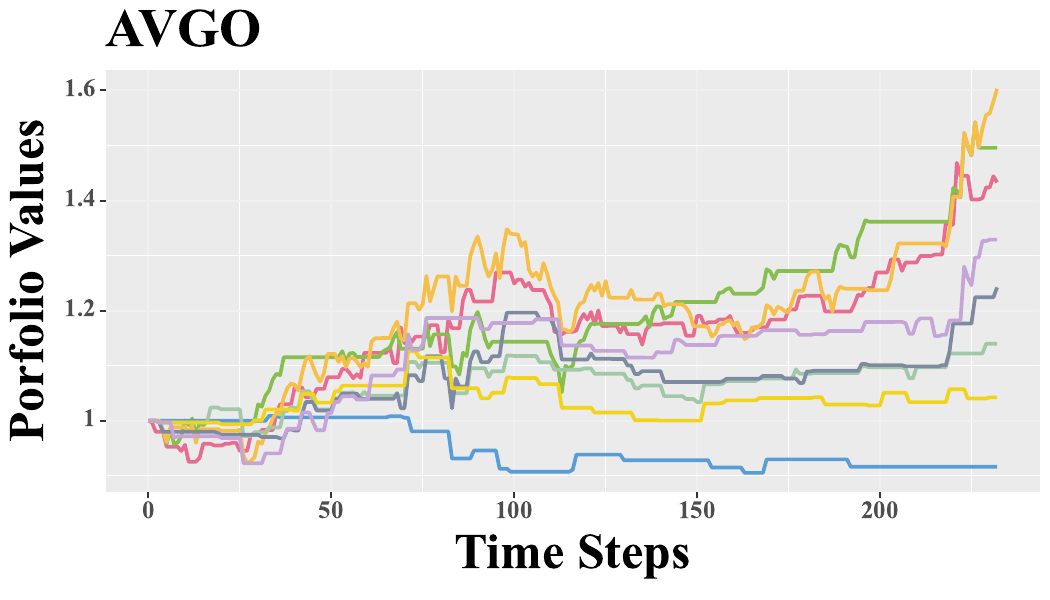}
		\label{fig:clr_yelp_light}
	\end{subfigure}
	\begin{subfigure}[t]{0.24\linewidth}
		\includegraphics[width=\textwidth]{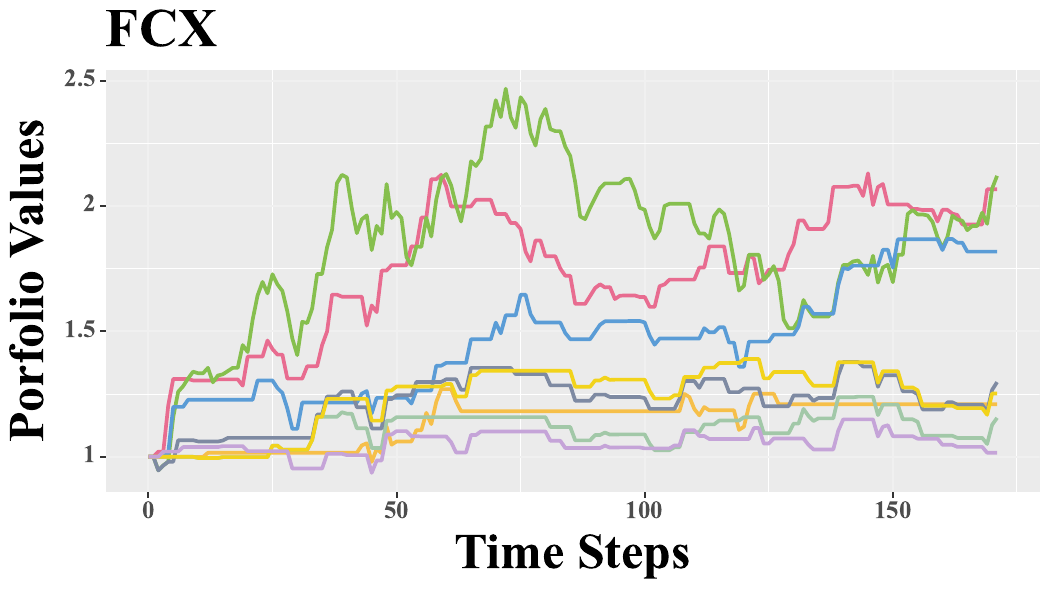}
		\label{fig:clr_yelp_wmf}
	\end{subfigure}
        \begin{subfigure}[t]{0.24\linewidth}
		\includegraphics[width=\textwidth]{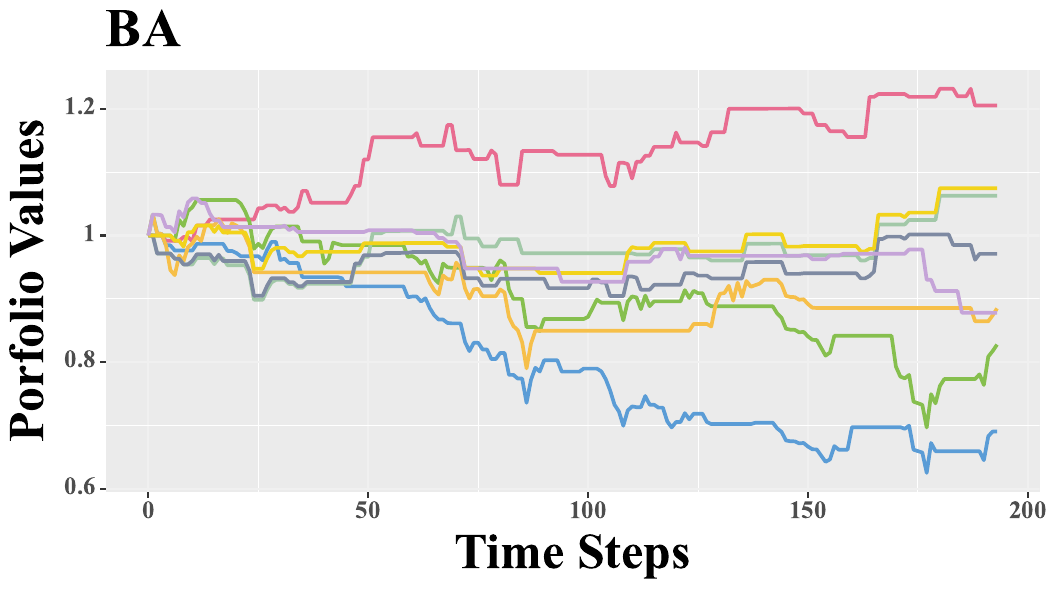}
		\label{fig:clr_ml_light}
	\end{subfigure}
	\begin{subfigure}[t]{0.24\linewidth}
		\includegraphics[width=\textwidth]{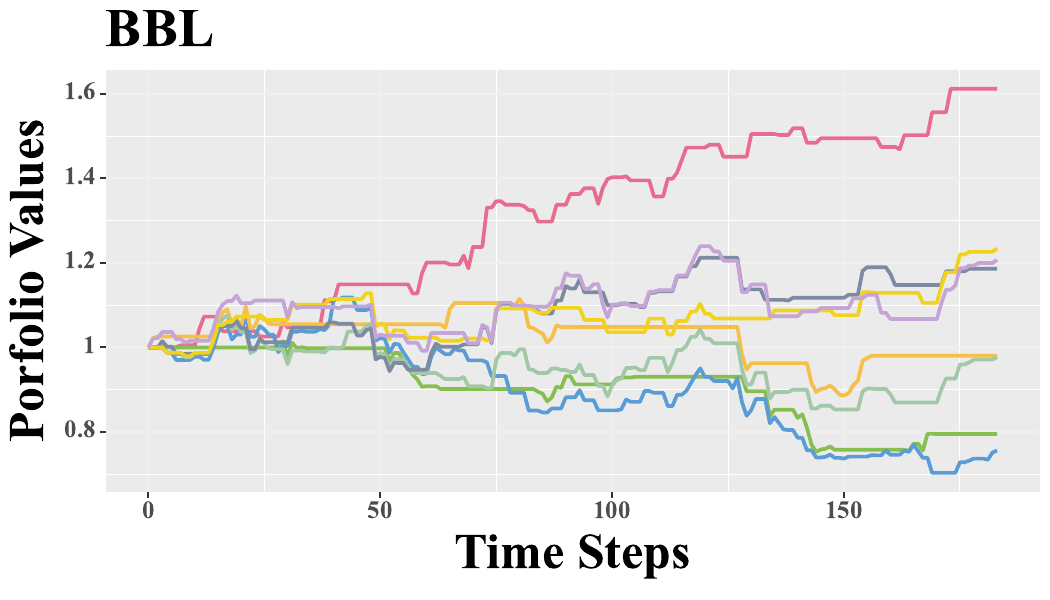}
		\label{fig:clr_ml_wmf}
	\end{subfigure}
	\begin{subfigure}[t]{0.24\linewidth}
		\includegraphics[width=\textwidth]{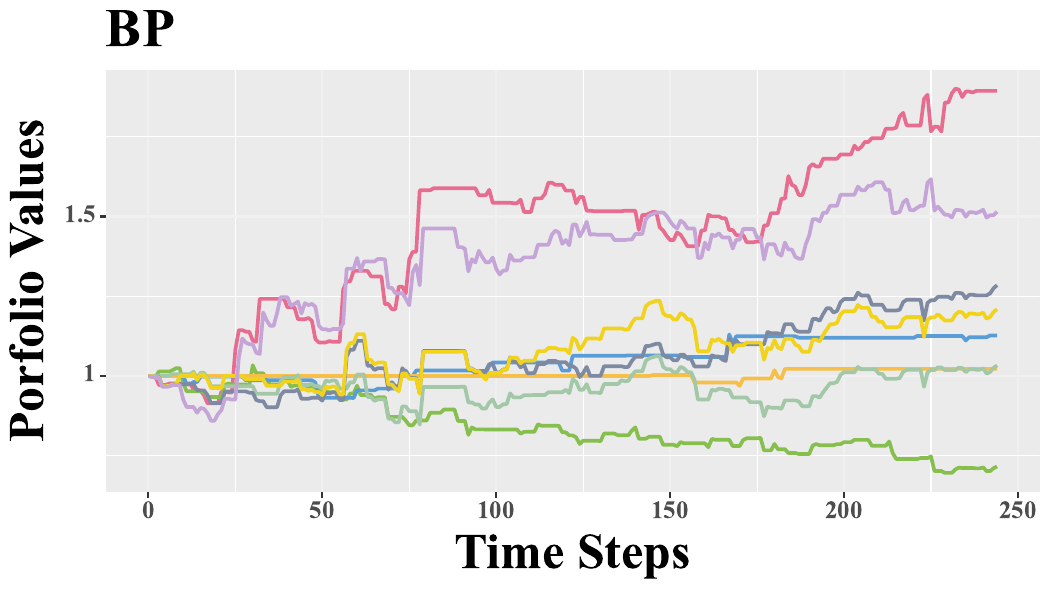}
		\label{fig:clr_yelp_light}
	\end{subfigure}
	\begin{subfigure}[t]{0.24\linewidth}
		\includegraphics[width=\textwidth]{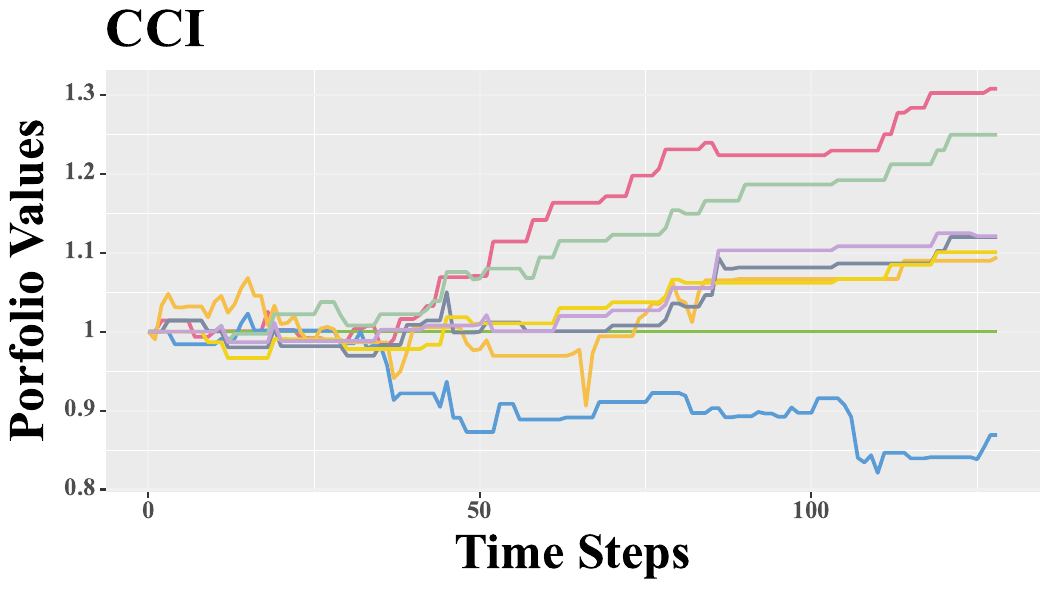}
		\label{fig:clr_yelp_wmf}
	\end{subfigure}

	\begin{subfigure}[t]{0.24\linewidth}
		\includegraphics[width=\textwidth]{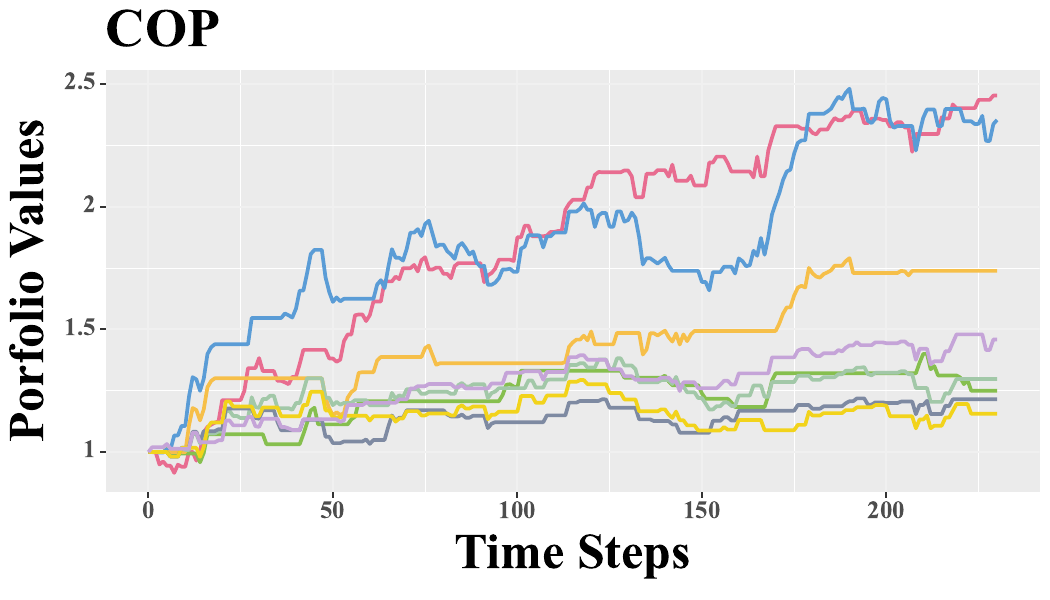}
		\label{fig:clr_ml_light}
	\end{subfigure}
	\begin{subfigure}[t]{0.24\linewidth}
		\includegraphics[width=\textwidth]{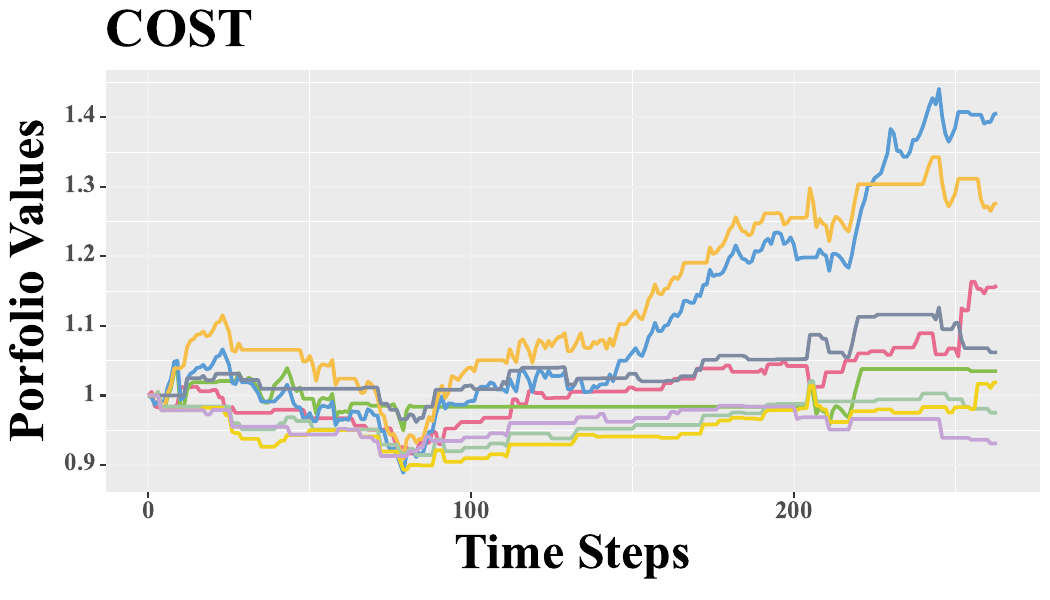}
		\label{fig:clr_ml_light}
	\end{subfigure}
	\begin{subfigure}[t]{0.24\linewidth}
		\includegraphics[width=\textwidth]{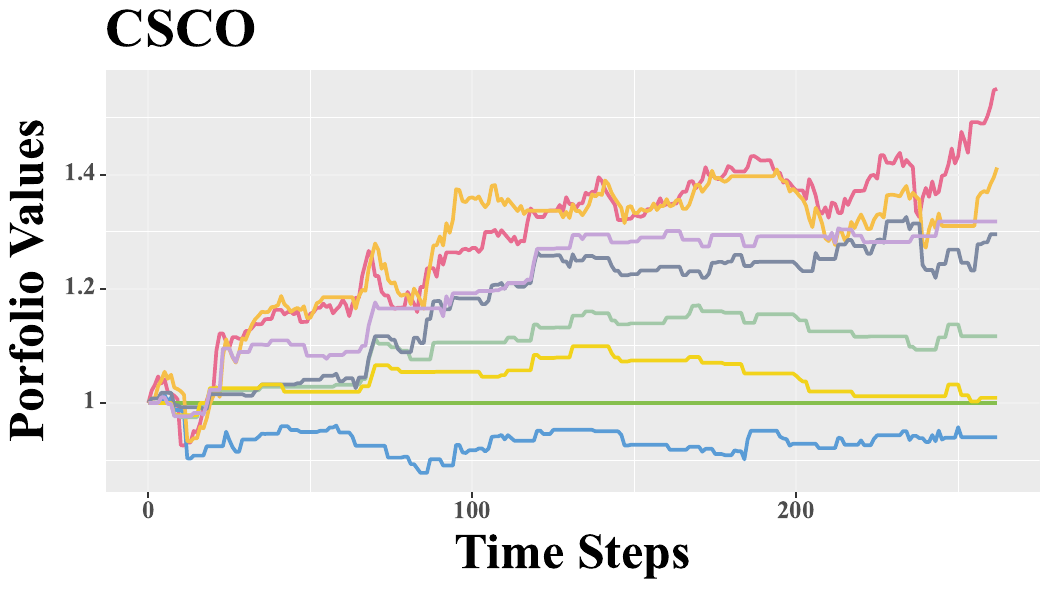}
		\label{fig:clr_ml_wmf}
	\end{subfigure}
	\begin{subfigure}[t]{0.24\linewidth}
		\includegraphics[width=\textwidth]{picture/lineplot_new/CVX.pdf}
		\label{fig:clr_yelp_light}
	\end{subfigure}
	\begin{subfigure}[t]{0.24\linewidth}
		\includegraphics[width=\textwidth]{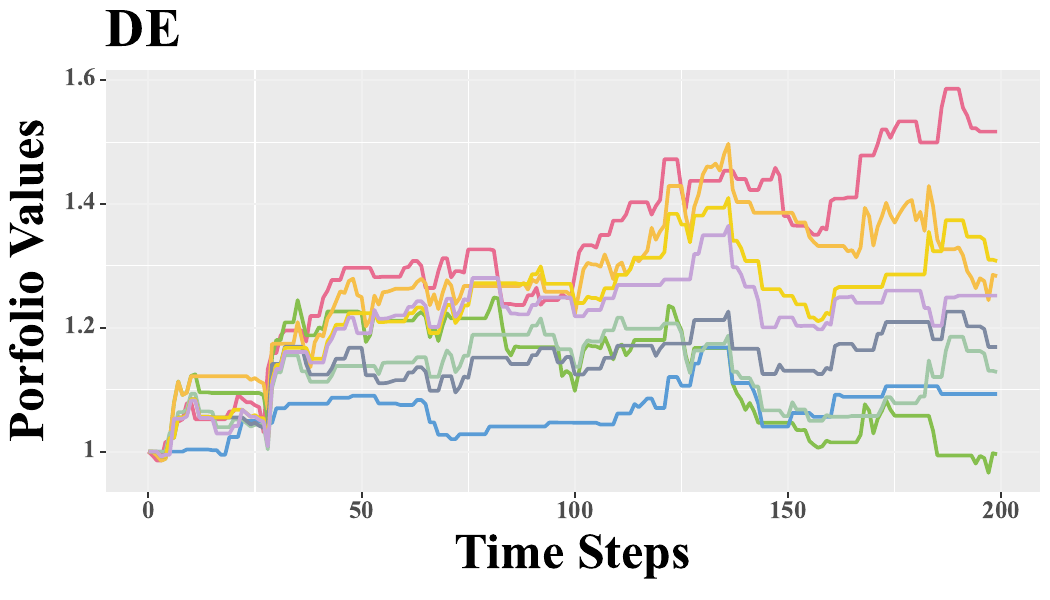}
		\label{fig:clr_yelp_wmf}
	\end{subfigure}
	\begin{subfigure}[t]{0.24\linewidth}
		\includegraphics[width=\textwidth]{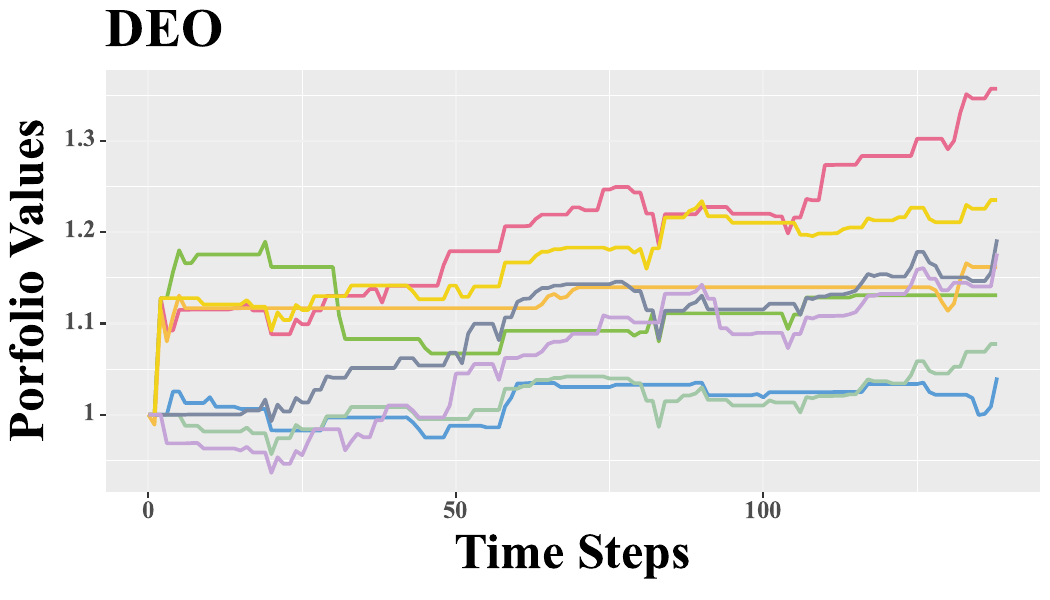}
		\label{fig:clr_ml_wmf}
	\end{subfigure}
	\begin{subfigure}[t]{0.24\linewidth}
		\includegraphics[width=\textwidth]{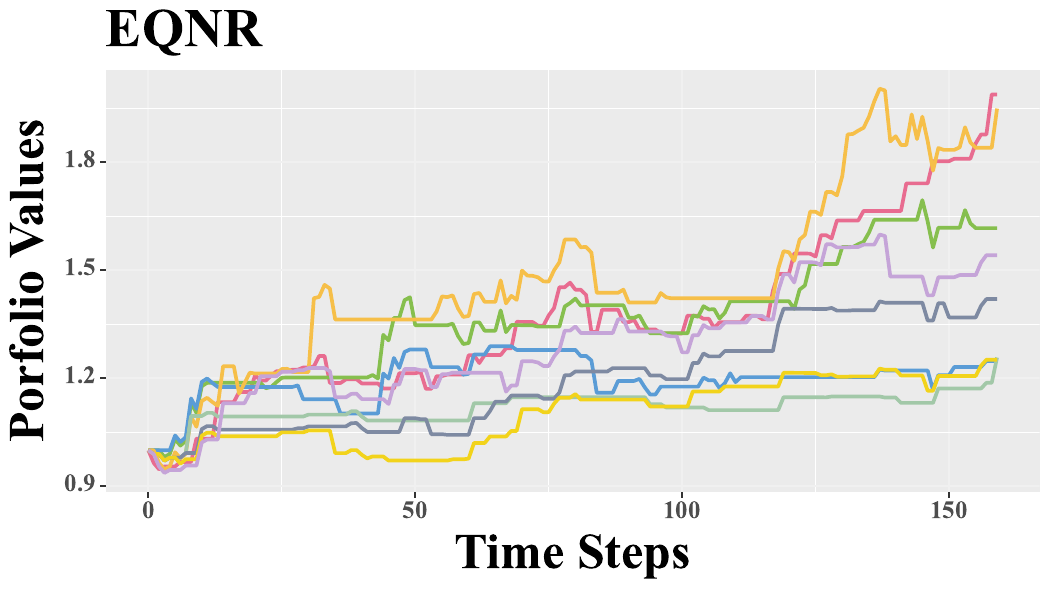}
		\label{fig:clr_yelp_light}
	\end{subfigure}
	\begin{subfigure}[t]{0.24\linewidth}
		\includegraphics[width=\textwidth]{picture/lineplot_new/EQNR.pdf}
		\label{fig:clr_yelp_wmf}
	\end{subfigure}

	\caption{
         The comparative performance of eight distinct models on different stock over a series of time steps. The models include \ours, LSTM, Stocknet, Transformer, CE-DNN, and ESPMP, along with DUAL-DNN and SCL-DNN highlighted in the legend. Each line represents the progression of portfolio values.
	}
 \label{fig:comparison2}
\end{figure*}

%% file: tables/appendix_industry_compare.tex
\begin{table*}[!htbp]
\centering
\caption{Comparative performances across industries. The table presents the Cumulative Returns, Annual Returns, Max Drawdown, and Calmar Ratio for each model, including \ours, LSTM, StockNet, Transformer, ESPMP, DUAL-DNN, and SCL-DNN. We stylize the best result as \textbf{bold} and the second best as \underline{underline}. Note: For Max Drawdown, the smaller the better.}
\label{tab:industry_compare}
\scalebox{.83}{
    \begin{tabular}{l|l|c|c|c|c|c|c|c|c|c}
    \toprule
    \midrule
      \textbf{Industry} & \textbf{Method} & \textbf{\ours~}&   \textbf{LSTM} & \textbf{Transformer} & \textbf{StockNet} & \textbf{DUALformer} & \textbf{CEformer} & \textbf{SCLformer} &  \textbf{ESPMP} & \textbf{\emph{Improve(\%)}} \\
      \midrule
      I1 & Cumulative Returns \textuparrow &    \textbf{0.193} &  0.046 &       0.074 &    0.053 &    0.017 &  0.009 &  -0.024 &  \underline{0.083} & +132.53 \\
              & Annual Return \textuparrow &    \textbf{0.215} &  0.069 &       0.079 &    0.082 &    0.016 &  0.019 &  -0.023 &  \underline{0.107} & +100.93 \\
              & Max Drawdown \textdownarrow &    \textbf{0.038} &  0.100 &       0.164 &    \underline{0.062} &    0.140 &  0.106 &   0.113 &  0.102 & +63.16 \\
              & Calmar Ratio \textuparrow &    \textbf{2.189} &  \underline{1.823} &       0.705 &    1.306 &    0.245 &  0.216 &  -0.237 &  1.043 & +20.08 \\
                 \midrule
      I2 & Cumulative Returns \textuparrow &    \textbf{0.185} & -0.001 &       0.024 &    \underline{0.040} &    0.032 & -0.049 &  -0.026 & -0.022 & +362.50 \\
              & Annual Return \textuparrow &    \textbf{0.223} & -0.018 &       0.010 &    0.030 &    \underline{0.092} & -0.045 &  -0.016 &  0.014 & +142.39 \\
              & Max Drawdown \textdownarrow &    \textbf{0.011} &  0.147 &       \underline{0.116} &    0.176 &    0.133 &  0.156 &   0.178 &  0.136  & +954.54 \\
              & Calmar Ratio \textuparrow &    \underline{2.637} & -1.068 &      -0.005 &    0.081 &    \textbf{4.969} &  0.594 &   0.705 &  2.034 & -46.93 \\
                 \midrule
      I3 & Cumulative Returns \textuparrow &    \textbf{0.237} &  0.064 &       0.103 &    0.078 &    0.053 &  \underline{0.107} &   0.103 &  0.065 & +121.50 \\
              & Annual Return \textuparrow &    \textbf{0.317} &  0.066 &       0.122 &    0.092 &    0.082 &  0.147 &   \underline{0.149} &  0.089 & +112.75 \\
              & Max Drawdown \textdownarrow &    \textbf{0.033} &  0.129 &       0.115 &    0.072 &    0.065 &  0.068 &   \underline{0.057} &  0.076 & +72.72 \\
              & Calmar Ratio \textuparrow &    \textbf{5.118} &  1.828 &       2.804 &    1.760 &    1.723 &  2.613 &   \underline{3.483} &  1.415 & +46.94 \\
                 \midrule
      I4 & Cumulative Returns \textuparrow &    \textbf{0.688} &  \underline{0.307} &       0.269 &    0.229 &    0.272 &  0.248 &   0.225 &  0.264 & +124.10 \\
              & Annual Return \textuparrow &    \textbf{0.847} &  0.306 &       0.311 &    0.231 &    0.300 &  0.277 &   0.233 &  \underline{0.321} & +163.86 \\
              & Max Drawdown \textdownarrow &    \textbf{0.067} &  0.202 &       0.241 &    0.165 &    0.138 &  \underline{0.134} &   0.156 &  0.159 & +100.00 \\
              & Calmar Ratio \textuparrow &    \textbf{4.8794} &  1.891 &       2.218 &    1.317 &   1.971 &  \underline{3.452} &   2.352 &  2.603 & +41.31 \\
                 \midrule
      I5 & Cumulative Returns \textuparrow &    \textbf{0.432} &  0.084 &       \underline{0.323} &    0.083 &    0.112 &  0.114 &   0.112 &  0.120 & +33.75 \\
              & Annual Return \textuparrow &    \textbf{0.648} &  0.107 &       \underline{0.342} &    0.102 &    0.245 &  0.255 &   0.301 &  0.164 & +89.47 \\
              & Max Drawdown \textdownarrow &    \textbf{0.074} &  0.122 &       0.123 &    0.121 &    \underline{0.081} &  0.085 &   0.092 &  0.115 & +9.46 \\
              & Calmar Ratio \textuparrow &   \underline{27.735} &  1.135 &       2.770 &    1.033 &    6.379 &  9.778 &  \textbf{40.917} &  3.317 & -32.22 \\
                 \midrule
      I6 & Cumulative Returns \textuparrow &    \textbf{0.417} &  0.038 &       0.061 &    \underline{0.210} &    0.035 &  0.101 &   0.095 &  0.130 & +98.57 \\
              & Annual Return \textuparrow &    \textbf{0.615} &  0.041 &       0.046 &    \underline{0.337} &    0.080 &  0.158 &   0.152 &  0.185 & +81.60 \\
              & Max Drawdown \textdownarrow &    \textbf{0.018} &  \underline{0.083} &       0.101 &    0.116 &    0.103 &  0.094 &   0.090 &  0.090 & +361.11 \\
              & Calmar Ratio \textuparrow &    \textbf{7.126} &    2.111 &       1.042 &    2.148 &    1.245 &  1.769 &   1.949 &  \underline{2.253} & +216.29 \\
                 \midrule
      I7 & Cumulative Returns \textuparrow &    \textbf{0.296} & -0.013 &       0.061 &    0.035 &    0.039 &  0.006 &   \underline{0.082} &  0.048 & +260.98 \\
              & Annual Return \textuparrow &    \textbf{0.328} & -0.020 &       0.066 &    0.026 &    0.042 &  0.005 &   \underline{0.092} &  0.049 & +256.52 \\
              & Max Drawdown \textdownarrow &    \textbf{0.033} &  \underline{0.129} &       0.196 &    0.146 &    0.140 &  0.143 &   0.137 &  0.179 & +290.91 \\
              & Calmar Ratio \textuparrow &    \textbf{2.477} &    0.128 &       0.282 &    \underline{1.133} &    0.409 &  0.400 &   0.724 &  0.466 & +118.62 \\
                 \midrule
      I8 & Cumulative Returns \textuparrow &    \textbf{0.416} &  0.167 &       0.072 &    \underline{0.285} &    0.116 &  0.122 &   0.094 &  0.069 & +45.96 \\
              & Annual Return \textuparrow &    \textbf{0.758} &  \underline{0.292} &       0.228 &    0.607 &    0.201 &  0.212 &   0.164 &  0.321 & +159.59 \\
              & Max Drawdown \textdownarrow &    \textbf{0.067} &  0.202 &       0.276 &    0.391 &    0.138 &  \underline{0.134} &   0.156 &  0.159 & +100.00 \\
              & Calmar Ratio \textuparrow &    \textbf{5.695} &  \underline{4.990} &       2.263 &    4.757 &    3.323 &  2.615 &   2.632 &  2.214 & +14.12 \\
                 \midrule
      I9 & Cumulative Returns \textuparrow &    \textbf{0.301} & -0.013 &       \underline{0.161} &    0.123 &    0.097 &  0.086 &   0.095 &  0.153 & +86.96 \\
              & Annual Return \textuparrow &    \textbf{0.729} & -0.001 &       \underline{0.490} &    0.317 &    0.216 &  0.207 &   0.172 &  0.297 & +48.78 \\
              & Max Drawdown \textdownarrow &    \textbf{0.064} &  0.125 &       0.124 &    0.087 &    0.090 &  0.087 &   0.076 &  \underline{0.064} & +0 \\
              & Calmar Ratio \textuparrow &   \textbf{12.515} &  0.438 &       4.154 &    \underline{6.336} &    3.915 &  5.902 &   3.143 &  5.990 & +185.22 \\
                 \midrule
      I10 & Cumulative Returns \textuparrow &    \textbf{0.416} &  0.025 &       0.090 &    0.110 &    0.077 &  0.131 &   0.066 &  \underline{0.202} & +105.94 \\
              & Annual Return \textuparrow &    \textbf{0.430} &  0.027 &       0.093 &    0.173 &    0.081 &  0.125 &   0.069 &  \underline{0.187} & +129.95 \\
              & Max Drawdown \textdownarrow &    \textbf{0.032} &  \underline{0.102} &       0.145 &    0.141 &    0.111 &  0.112 &   0.112 &  0.108 & +218.75 \\
              & Calmar Ratio \textuparrow &    \textbf{4.173} &  2.113 &       1.135 &    2.059 &    0.857 &  1.413 &   1.357 &  \underline{2.532} & +64.81 \\
    \midrule
    \bottomrule
    \end{tabular}
    }
\end{table*}